%% file: main.tex
\documentclass[10pt,twocolumn,letterpaper]{article}

\usepackage[pagenumbers]{cvpr} % To force page numbers, e.g. for an arXiv version

\input{preamble}
\definecolor{cvprblue}{rgb}{0.21,0.49,0.74}
\usepackage[pagebackref,breaklinks,colorlinks,allcolors=cvprblue]{hyperref}
\usepackage{pifont}

\def\confName{CVPR}
\def\confYear{2027}

\title{Learning Dense Head-Object Contact Estimation with Synthetic Data}

\author{
  Daniel Sungho Jung$^{1}$ \hskip1.6em Kyoung Mu Lee$^{1,2}$ \\
   $^{1}$IPAI, $^{2}$Dept. of ECE \& ASRI, Seoul National University, Korea  \\ 
   {\tt\small \{dqj5182, kyoungmu\}@snu.ac.kr} 
}

\begin{document}
\maketitle
\input{sec/0_abstract}    
\input{sec/1_introduction}
\input{sec/2_related_works}
\input{sec/3_proposed_method}
\input{sec/4_dataset}
\input{sec/5_implementation_details}
\input{sec/6_experiments}
\input{sec/7_conclusion}
\input{sec/X_suppl}

\clearpage

\input{sec/8_llm_statement}

{
    \small
    \bibliographystyle{ieeenat_fullname}
    \bibliography{main}
}

\end{document}

%% file: sec/0_abstract.tex
\begin{abstract}
Head serves a critical role in human cognitive and perceptual functions while remaining vulnerable to external impacts, making the understanding of head contact essential for human safety.
Despite its importance, dense head-object contact estimation remains largely unexplored, primarily due to the absence of head-object contact data.
Recently, it has been shown that synthetic data can serve as powerful training data in limited data scenario, implying the potential of synthetic data in head-object contact estimation.
There are two core challenges for dense head-object contact estimation.
First, head-object interactions are highly diverse, resulting in substantial variations in contact regions and patterns across different objects and interaction types.
Second, head contact regions are often heavily occluded by hair, making direct visual observation of the actual contact regions difficult.
To address these issues, we introduce HEad COntact estimation (HECO) with synthetic data.
To handle diverse head-object interactions, we leverage semantic context from text features that encode semantic relationships between head regions and interacting objects.
To resolve hair occlusion issue of actual head contact regions, we introduce occlusion-aware contact learning that learns to infer head contact beyond visible information from images via amodal inpainting.
As a result, we enable head-object contact estimation with HECO trained with our synthetic dataset, overcoming the issue of absence of contact data. 
The codes will be released.
\end{abstract}

%% file: sec/1_introduction.tex
\section{Introduction}

\input{fig/heco_dataset}

Humans perform fundamental cognitive functions such as thinking, perceiving, and reasoning with their heads, making the head one of the most important body regions.
Understanding head--object contact is therefore relevant to human safety and physical interaction analysis.
Impact location influences regional brain tissue deformation~\cite{elkin2019brain}, making contact localization useful for characterizing head impacts.
In sports, head--ball contact location helps characterize heading technique~\cite{shewchenko2005heading}.
These applications motivate dense head--object contact estimation, which predicts per-vertex contact on a 3D head mesh from a single RGB image to capture the location of head contact.

Despite its importance, there lacks 3D human-object interaction dataset that provides diverse head-object interaction both in 3D geometry and RGB image. 
There are two lines of datasets, capture-based datasets and image-based datasets, that have potential for providing useful head-object interaction data.
Capture-based datasets~\cite{taheri2020grab, li2023object, lv2024himo} provide detailed human--object interaction geometry, but offer limited paired RGB data with human and object appearance.
Image-based datasets~\cite{hassan2019resolving, bhatnagar2022behave, huang2022capturing, jiang2023full, zhang2023neuraldome, fan2023arctic, huang2024intercap} provide RGB images paired with 3D geometry, but primarily target general human--object or human--scene interactions, with very limited number of samples that contain head contact.
Consequently, existing datasets are insufficient for learning dense head--object contact estimation from RGB images, particularly for robustly handling diverse head regions and object categories.
To address this gap, we construct a dedicated head--object interaction dataset COHE that provides both RGB images and dense head contact annotations derived from accurate 3D head and object geometry.

There are two core challenges for dense head--object contact estimation.
First, contact patterns vary substantially across different head regions.
A cup rim contacts the lips during drinking, a phone is held against the ear, and headgear can contact extended regions of the scalp.
These interactions involve distinct head regions with substantially different contact locations and spatial extents, ranging from localized contact on small facial regions to broad contact across the head surface.
Such region-dependent variations make it challenging to accurately localize contact across the entire head, as different head regions exhibit distinct contact patterns depending on their interactions with surrounding objects.
Accurate localization therefore requires understanding the semantic relationships between head regions and interacting objects, while leveraging local visual evidence to examine the contact area.
Second, hair often occludes the underlying head surface on which contact occurs.
Hair alters the apparent head boundary, creating a discrepancy between the visible hair-covered head and the actual head surface.
This discrepancy is further compounded by occlusion from interacting objects, making visible overlap insufficient to determine the actual contact region.
Moreover, different hairstyles can produce substantially different appearances for similar underlying head--object configurations, introducing additional ambiguity in contact localization.
A robust estimator must therefore capture the semantic relationships governing diverse interactions and reason about the underlying head surface and head--object configuration beyond directly observable visual information.

Therefore, we propose HECO, a framework for dense head--object contact estimation.
Initially, we build a synthetic head--object interaction dataset, COHE, consisting of three data types: capture-based, physics-based, and cross-source.
Capture-based data leverages accurate 3D geometry from motion-capture human--object interaction datasets~\cite{taheri2020grab, li2023object, jiang2023full, zhang2023neuraldome, lv2024himo} and applies appearance transfer to generate textured human--object interactions.
Physics-based data utilizes a physics engine~\cite{Macklin_Warp_A_High-performance_2022} to simulate contact between textured 3D humans and objects across diverse head regions and interaction configurations.
Cross-source data combines the two data types by replacing objects from motion-capture interactions with novel 3D object assets through curvature-driven object replacement.
Together, these data types provide diverse head--object interactions with aligned RGB images and dense contact annotations.
Based on COHE dataset, we employ two strategies to address diverse head-region contact patterns and hair occlusion.
First, to handle diverse contact patterns across head regions, we introduce text-guided semantic alignment that leverages semantic context from a pretrained vision--language model~\cite{tschannen2025siglip}.
Specifically, we align image features with text features that encode the object category, contact state, and involved head region, encouraging the model to learn semantic relationships between head regions and interacting objects.
Second, to address hair occlusion, we introduce hair occlusion-aware learning that reconstructs both the hairless head without the interacting object and the hairless head with the object retained, along with amodal head and object masks.
This allows the model to reason about the underlying head surface and its spatial relationship with interacting objects beyond the visible hair-covered appearance.
Finally, the learned features are combined through spatial attention and decoded into dense contact predictions on the FLAME~\cite{li2017learning} head mesh.

As a result, HECO achieves state-of-the-art performance across diverse head--object interactions by leveraging semantic context and hair occlusion-aware learning for dense head contact estimation.

Our key contributions are as follows:
\begin{itemize}
\item We introduce HECO, a framework for dense head--object contact estimation from a single RGB image, along with COHE, a synthetic dataset that provides diverse head--object interactions with aligned RGB images and dense contact annotations.
\item To handle diverse contact patterns across head regions, we introduce text-guided semantic alignment that leverages semantic context from a pretrained vision--language model to learn semantic relationships between head regions and interacting objects.
\item To address hair occlusion, we propose hair occlusion-aware learning that reconstructs hairless head and head--object appearances along with amodal masks, enabling contact reasoning beyond visible head surfaces.
\item In the end, HECO demonstrates state-of-the-art performance in dense head--object contact estimation across diverse benchmarks, highlighting the effectiveness of semantic context and hair occlusion learning.
\end{itemize}

%% file: fig/heco_dataset.tex
\begin{figure*}[t]
\begin{center}
\includegraphics[width=1.0\linewidth]{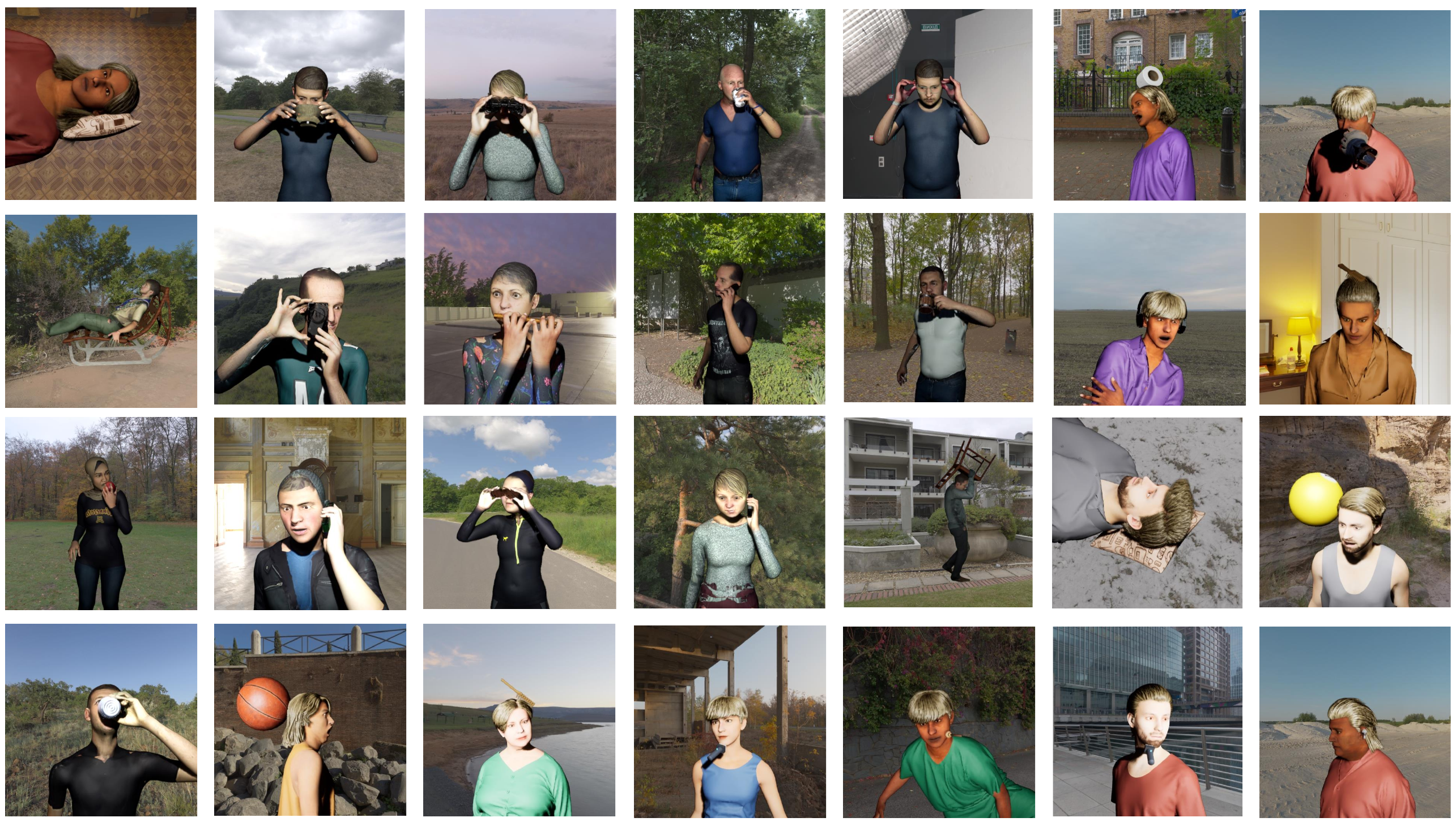}
\end{center}
\vspace{-4mm}
\caption{
\textbf{COHE Dataset.} We construct a dataset that contains diverse 3D head--object interactions with wide range of object categories.}
\label{fig:heco_dataset}
\vspace{-0.3cm}
\end{figure*}

%% file: sec/2_related_works.tex
\section{Related works}

\noindent\textbf{Dense head contact estimation.}
Head contact is highly underexplored with most of the existing works focus on body contact~\cite{hassan2021populating, huang2022capturing, tripathi2023deco, dwivedi2025interactvlm, sur2026unicon3r} or hand-face contact~\cite{wu2025dice}.
POSA~\cite{hassan2021populating} predicts dense body contact conditioned on body pose of SMPL-X~\cite{pavlakos2019expressive} human mesh via cVAE~\cite{sohn2015learning}.
BSTRO~\cite{huang2022capturing} introduces a BERT~\cite{devlin2019bert}-based dense per-vertex contact estimation framework and RICH dataset that annotates dense body contact.
DECO~\cite{tripathi2023deco} presents a dense human body contact estimation framework that leverages scene context from image segmentation and part context from part segmentation, along with DAMON dataset that enables large-scale dense contact annotation without motion capture.
InteractVLM~\cite{dwivedi2025interactvlm} utilizes pre-trained VLMs for 2D visual reasoning and performs multi-view 2D-to-3D lifting strategy for 3D contact reasoning.
DICE~\cite{wu2025dice} presents an end-to-end model for 3D hand-face interaction reconstruction that explicitly captures facial contact and deformation.
% UniCon3R~\cite{sur2026unicon3r} predicts 4D contact using scene context features, 3D world-frame pointmaps, and contact predictions from previous frames to incorporate scene geometry and temporal coherence.
Despite these advances, dense head-object contact estimation remains unexplored, motivating us to introduce HECO, a dense contact estimation framework, and COHE, a head-object contact dataset.

\noindent\textbf{Synthetic human dataset.}
Synthetic human data has been widely adopted to address the limited scale and diversity of real-world human data captured through motion capture.
SURREAL~\cite{varol2017learning} introduced a large-scale dataset with people rendered from 3D sequences of human motion capture data from CMU MoCap~\cite{cmu_mocap}.
RHD~\cite{zimmermann2017learning} created a synthetic dataset with various data augmentation options to compensate for lack of 3D hand pose data with sufficient variability.
SynthHands~\cite{mueller2017real} employed merged reality framework that tracks a real hand for 3D hand pose~\cite{sridhar2015fast} to interact with a virtual object, to build hand-object interaction dataset.
ObMan~\cite{hasson2019learning} presented a large-scale synthetic dataset of hand-object interactions using ShapeNet~\cite{chang2015shapenet} object meshes and Graspit~\cite{miller2004graspit} to generate plausible grasps.
PressurePose~\cite{clever2020bodies} built a synthetic dataset with 3D human meshes with pressure images generated by FleX~\cite{macklin2014unified} and DART~\cite{lee2018dart} physics simulator to drop 3D human model on a synthetic pressure-sensing mat.
% Ego3DHands~\cite{lin2021two}
AGORA~\cite{patel2021agora} obtained commercially-available, high-quality, textured human scans in diverse poses and natural clothing including children to build physically plausible and photorealistic synthetic dataset.
DARTset~\cite{gao2022dart} enlarged the diversity of 3D hand datasets by adding accessories and hand texture with different blemishes and make-ups.
BEDLAM~\cite{black2023bedlam} demonstrated that training solely on synthetic data can achieve state-of-the-art performance in 3D human pose and shape estimation.
Re:InterHand~\cite{moon2023dataset} and RenderIH~\cite{li2023renderih} addressed lack of variability of background and hand texture for 3D interacting hands.
WHAC-A-Mole~\cite{yin2024whac} tackled lack of data diversity on human motion and camera trajectory for world-grounded expressive human pose and shape estimation.
ProciGen~\cite{xie2024template} generated large-scale interaction dataset with diverse object shape by transferring dense semantic correspondence and contact.
GTA-Human~\cite{cai2024playing} generated large-scale 3D human dataset with the GTA-V game engine to build diverse set of subjects, actions, and scenarios.
HandSynthesis~\cite{zhao2025analyzing} analyzed the data gap between synthetic and real data for 3D hand pose estimation in terms of appearance, hand pose, object occlusion, and hand skeleton topology, and achieved performance comparable to real-data training using only synthetic data.
BEDLAM2.0~\cite{tesch2026bedlam2} extended BEDLAM by introducing a video dataset with more realistic and diverse human and camera motions, enabling human pose estimation in world coordinates.
Our COHE dataset tackles limited data issue of dense head contact by building three data types of capture-based, physics-based, and cross-source that effectively utilize motion-capture human-object interaction datasets~\cite{taheri2020grab, li2023object, jiang2023full, zhang2023neuraldome, lv2024himo}, physics engine~\cite{Macklin_Warp_A_High-performance_2022}, diverse human motion~\cite{tesch2026bedlam2} and object assets~\cite{deitke2023objaverse, downs2022google, tyree20226, stojanov2021using}.

%% file: sec/3_proposed_method.tex
\section{Method: HECO}
Figure~\ref{fig:pipeline} shows the overall pipeline of HECO that consists of synthetic-to-real augmentation, text-guided semantic alignment, hair occlusion-aware learning, occlusion attention, and head contact decoder.

\input{fig/heco_overall_pipeline}

\subsection{Synthetic-to-real augmentation}
Given an RGB image~$\mathbf{I} \in \mathbb{R}^{3 \times H \times W}$, where $H$ and $W$ represent the spatial resolution, we conduct amplitude spectrum augmentation~\cite{chattopadhyay2023pasta} on the image~$\mathbf{I}$ to enhance robustness to the synthetic-to-real domain gap.

\subsection{Text-guided semantic alignment}

From the image~$\mathbf{I}$, a SigLIP2~\cite{tschannen2025siglip} image encoder extracts an image feature~$\mathbf{F}_{\mathrm{i}} \in \mathbb{R}^{c \times h \times w}$ that possesses general visual-text alignment, where $c=1536$.
We also encode a text prompt~$\mathbf{T}$ that contains contact semantics following the template
\texttt{``An image of \{object\} \{contact state\} with the \{head part in contact\}``} with a frozen SigLIP2 text encoder to obtain a text feature~$\mathbf{F}_{\mathrm{t}}$.
A semantic adapter~$\mathbf{A}_{\mathrm{sem}}$, which consists of Layer Normalization~\cite{ba2016layer} followed by a two-layer MLP, refines $\mathbf{F}_{\mathrm{i}}$ to produce a contact semantics aligned image feature~$\mathbf{F}_{\mathrm{i}}^{\mathrm{sem}}$. 
We then aggregate the aligned image feature~$\mathbf{F}_{\mathrm{i}}^{\mathrm{sem}}$ by global average pooling~(GAP) to obtain the pooled image feature~$\mathbf{F}_{\mathrm{i,p}}^{\mathrm{sem}}$.
From the pooled image feature~$\mathbf{F}_{\mathrm{i,p}}^{\mathrm{sem}}$, we construct a normalized image semantic latent~$\mathbf{z}_{\mathrm{i}}$ as:
\begin{equation}
\mathbf{z}_{\mathrm{i}}
= \frac{\mathbf{F}_{\mathrm{i,p}}^{\mathrm{sem}}}
{\left\lVert \mathbf{F}_{\mathrm{i,p}}^{\mathrm{sem}} \right\rVert_2},
\qquad
\mathbf{F}_{\mathrm{i,p}}^{\mathrm{sem}}
= \frac{1}{hw}\sum_{q=1}^{hw}
  \mathbf{A}_{\mathrm{sem}}(\mathbf{F}_{\mathrm{i, q}}).
\end{equation}
Similarly, we construct the text semantic latent~$\mathbf{z}_{\mathrm{t}}$ from the pooled text feature~$\mathbf{F}_{\mathrm{t}}$ in the shared semantic space as follows:
\begin{equation}
    \mathbf{z}_{\mathrm{t}}
    = \frac{\mathbf{F}_{\mathrm{t}}}
    {\left\lVert \mathbf{F}_{\mathrm{t}} \right\rVert_2}.
\end{equation}
We then align the image semantic latent~$\mathbf{z}_{\mathrm{i}}$ with the text semantic latent~$\mathbf{z}_{\mathrm{t}}$.
The frozen text representation provides a stable semantic target that encourages the visual representation to retain cues about head contact.
The text guidance is used only during training and therefore requires no text input at inference time.

\subsection{Hair occlusion-aware learning}

From the semantics-aligned image feature~$\mathbf{F}_{\mathrm{i}}^{\mathrm{sem}} \in \mathbb{R}^{c \times h \times w}$, we obtain a spatial feature map~$\mathbf{F}_{\mathrm{s}} \in \mathbb{R}^{d \times h \times w}$ by applying a linear projection at each spatial location to change the feature dimension from $c$ to $d$, where $d=512$.
A segmentation decoder predicts head--object segmentation from~$\mathbf{F}_{\mathrm{s}}$, consisting of a visible head probability map~$\mathbf{P}_{\mathrm{h}}^{\mathrm{vis}}$ and a visible object probability map~$\mathbf{P}_{\mathrm{o}}^{\mathrm{vis}}$.
The predicted maps $\mathbf{P}_{\mathrm{h}}^{\mathrm{vis}}$ and $\mathbf{P}_{\mathrm{o}}^{\mathrm{vis}}$, which are analogous to holes in LaMa~\cite{suvorov2022resolution}, are used to construct two task-specific feature streams that build head feature~$\mathbf{F}_{\mathrm{h}}$ and head-object feature~$\mathbf{F}_{\mathrm{ho}}$.
Head gating suppresses only the visible head region while preserving the object, whereas head--object gating suppresses the union of the visible head and object regions as:
\begin{equation}
\begin{gathered}
\widetilde{\mathbf{F}}_{\mathrm{h}}
= \mathbf{F}_{\mathrm{s}} \odot (1-\mathbf{P}_{\mathrm{h}}^{\mathrm{vis}}), \\
\widetilde{\mathbf{F}}_{\mathrm{ho}}
= \mathbf{F}_{\mathrm{s}} \odot (1-\mathbf{P}_{\mathrm{h}}^{\mathrm{vis}})
\odot (1-\mathbf{P}_{\mathrm{o}}^{\mathrm{vis}}).
\end{gathered}
\end{equation}
where $\odot$ denotes element-wise multiplication.
The two gated feature streams are processed by independent Fast Fourier Convolution~(FFC) modules~\cite{suvorov2022resolution}, which combine local convolutional processing with global Fourier-domain reasoning to complete the suppressed regions in feature space.
Each completed feature is then passed to an independent DPT decoder~\cite{ranftl2021vision} to produce a head feature~$\mathbf{F}_{\mathrm{h}} \in \mathbb{R}^{256 \times 8h \times 8w}$ and a head--object feature~$\mathbf{F}_{\mathrm{ho}} \in \mathbb{R}^{256 \times 8h \times 8w}$:
\begin{equation}
\begin{gathered}
\mathbf{F}_{\mathrm{h}}
= \mathrm{DPT}_{\mathrm{h}}\!\left(
\mathrm{FFC}_{\mathrm{h}}(\widetilde{\mathbf{F}}_{\mathrm{h}})
\right), \\
\mathbf{F}_{\mathrm{ho}}
= \mathrm{DPT}_{\mathrm{ho}}\!\left(
\mathrm{FFC}_{\mathrm{ho}}(\widetilde{\mathbf{F}}_{\mathrm{ho}})
\right).
\end{gathered}
\end{equation}
Finally, the head--object feature~$\mathbf{F}_{\mathrm{ho}}$ is fed to a head image decoder and a head mask decoder to reconstruct the hairless head image~${\mathbf{I}}_{\mathrm{h}}$ and estimate the amodal head mask~${\mathbf{M}}_{\mathrm{h}}$, respectively.
In parallel, the head feature~$\mathbf{F}_{\mathrm{h}}$ is fed to a head--object image decoder and an object mask decoder to reconstruct the hairless head--object image~${\mathbf{I}}_{\mathrm{ho}}$ and estimate the amodal object mask~${\mathbf{M}}_{\mathrm{o}}$.
This design allows the head--object-gated stream to remove both the hair and the interacting object for hairless head reconstruction, while the head-gated stream removes the hair but retains the object for hairless head--object reconstruction.

\subsection{Occlusion attention}
We fuse the head feature~$\mathbf{F}_{\mathrm{h}}$ and head--object feature~$\mathbf{F}_{\mathrm{ho}}$ using a spatial attention module.
We first apply adaptive average pooling to reduce the spatial resolution from $8h\times8w$ to $h\times w$, obtaining $\mathbf{F}_{\mathrm{h}}^{\mathrm{p}},\mathbf{F}_{\mathrm{ho}}^{\mathrm{p}}\in\mathbb{R}^{256\times h\times w}$.
The pooled features are concatenated and passed through a $3\times3$ convolution, ReLU~\cite{nair2010rectified}, and dropout~\cite{srivastava2014dropout} with a rate of $0.2$, followed by a $1\times1$ convolution that produces two logits.
The logits are normalized with a channel-wise softmax with temperature $\tau=1.5$ to produce spatial weights~$w_{\mathrm{h}},w_{\mathrm{ho}}\in\mathbb{R}^{1\times h\times w}$.
The fused feature is then computed as:
\begin{equation}
\mathbf{F}_{\mathrm{fuse}}
=w_{\mathrm{h}}\odot\mathbf{F}_{\mathrm{h}}^{\mathrm{p}}
+w_{\mathrm{ho}}\odot\mathbf{F}_{\mathrm{ho}}^{\mathrm{p}},
\end{equation}
where $\odot$ denotes element-wise multiplication.
Finally, a contact adapter~$\mathbf{A}_{\mathrm{c}}$ projects the concatenation of the pooled features~$\mathbf{F}_{\mathrm{h}}^{\mathrm{p}}$,~$\mathbf{F}_{\mathrm{ho}}^{\mathrm{p}}$ and the fused feature~$\mathbf{F}_{\mathrm{fuse}}$, and its output is added to the spatial feature map~$\mathbf{F}_{\mathrm{s}}$ to produce the contact feature~$\mathbf{F}_{\mathrm{c}}\in\mathbb{R}^{d\times h\times w}$ as:
\begin{equation}
\mathbf{F}_{\mathrm{c}}
=\mathbf{F}_{\mathrm{s}}+\mathbf{A}_{\mathrm{c}}\!\left(
\mathrm{Concat}\!\left(
\mathbf{F}_{\mathrm{h}}^{\mathrm{p}},
\mathbf{F}_{\mathrm{ho}}^{\mathrm{p}},
\mathbf{F}_{\mathrm{fuse}}
\right)\right),
\end{equation}
where $\mathrm{Concat}$ denotes channel-wise concatenation.
This spatial attention fusion adaptively combines the head feature~$\mathbf{F}_{\mathrm{h}}$ and head--object feature~$\mathbf{F}_{\mathrm{ho}}$ for constructing the contact feature~$\mathbf{F}_{\mathrm{c}}$.

\subsection{Head contact decoder}
Following HACO~\cite{jung2025learning}, we introduce learnable contact embeddings~$\mathbf{E}_{\mathrm{c}} \in \mathbb{R}^{V \times d}$ as contact query tokens, with one token assigned to each of the $V=5{,}023$ FLAME vertices.
To encode the FLAME topology on the query token, we add a linear projection of FLAME canonical vertex coordinates to the corresponding contact embedding~$\mathbf{E}_{\mathrm{c}}$ as follows:
\begin{equation}
\mathbf{Q}^{(0)}
= \mathbf{E}_{\mathrm{c}} + \phi(\mathbf{X}_{\mathrm{FLAME}}),
\end{equation}
where $\mathbf{Q}^{(0)} \in \mathbb{R}^{V \times d}$ denotes the initialized contact query tokens, $\mathbf{X}_{\mathrm{FLAME}} \in \mathbb{R}^{V \times 3}$ contains the canonical FLAME vertex coordinates, and $\phi:\mathbb{R}^{3}\rightarrow\mathbb{R}^{d}$ is a linear projection.
Given the contact feature~$\mathbf{F}_{\mathrm{c}} \in \mathbb{R}^{d \times h \times w}$, we flatten it into spatial tokens and prepend a global token obtained by average pooling, resulting in the visual context~$\overline{\mathbf{F}}_{\mathrm{c}} \in \mathbb{R}^{(hw+1) \times d}$.
Using~$\mathbf{Q}^{(0)}$ as queries and~$\overline{\mathbf{F}}_{\mathrm{c}}$ as keys and values, the contact decoder applies four Transformer~\cite{vaswani2017attention} layers, each comprising eight-head cross-attention and a feed-forward network, and maps the resulting query tokens to vertex-level contact logits using a linear classifier:
\begin{equation}
\mathbf{C}
= \mathrm{Dec}_{\mathrm{c}}\!\left(
\mathbf{Q}^{(0)},\overline{\mathbf{F}}_{\mathrm{c}}
\right),
\qquad
\mathbf{C}\in\mathbb{R}^{V}.
\end{equation}
This results in vertex-level contact logits~$\mathbf{C}\in\mathbb{R}^{V}$ for the FLAME vertices.

\subsection{Final outputs and loss functions}

\noindent\textbf{Multi-level outputs.}
Head contact decoder produces vertex-level logits $\mathbf{C}\in\mathbb{R}^{V}$ on the full FLAME~\cite{li2017learning} head mesh, where $V=5{,}023$. 
Following the multi-level supervision strategy of HACO~\cite{jung2025learning}, we construct $N=5$ fixed regressors based on the CoMA~\cite{ranjan2018generating} defined FLAME mesh hierarchy, where $i=1,\ldots,N$ indexes the hierarchy levels and $v_i\in\{5{,}023,1{,}256,314,79,20\}$ denotes the number of vertices at level $i$. 
At each coarse level, we use the vertices retained in the CoMA hierarchy as representative vertices for constructing the corresponding regressor. 
Each full-resolution vertex is then assigned to its geodesically nearest representative vertex at that level. 
This assignment defines a fixed regressor $\mathcal{R}_i^{v_i}\in\mathbb{R}^{v_i\times V}$ that averages the vertices assigned to each representative vertex. 
The multi-level logits are obtained as:
\begin{equation}
\mathbf{C}_i=\mathcal{R}_i^{v_i}\mathbf{C},
\qquad i=1,\ldots,N.
\label{eq:heco_multi_level_regressor}
\end{equation}
Contact probabilities at each level are then computed by applying a sigmoid after projection as:
\begin{equation}
\hat{\mathbf{C}}_i=\sigma(\mathbf{C}_i),
\qquad i=1,\ldots,N.
\end{equation}
The coarse outputs are derived solely based on the regressors and thus do not require separate prediction heads.

\noindent\textbf{Loss functions.}
We train HECO in an end-to-end manner using the following objective:
\begin{equation}
\begin{aligned}
\mathcal{L} ={}&
\lambda_{\mathrm{contact}}\mathcal{L}_{\mathrm{contact}}
+ \lambda_{\mathrm{sem\text{-}align}}\mathcal{L}_{\mathrm{sem\text{-}align}} \\
&+ \lambda_{\mathrm{segm}}\mathcal{L}_{\mathrm{segm}}
+ \lambda_{\mathrm{img}}\mathcal{L}_{\mathrm{img}}
+ \lambda_{\mathrm{mask}}\mathcal{L}_{\mathrm{mask}}.
\end{aligned}
\end{equation}
The contact loss~$\mathcal{L}_{\mathrm{contact}}$ is the sum of contact losses computed at five levels of the FLAME~\cite{li2017learning} mesh hierarchy defined by CoMA~\cite{ranjan2018generating}, with $5{,}023$, $1{,}256$, $314$, $79$, and $20$ vertices at each level.
The contact loss~$\mathcal{L}_{\mathrm{contact}}$ combines weighted binary cross-entropy loss and Dice loss~\cite{milletari2016v} to mitigate class imbalance between contact and non-contact regions and encourage accurate overlap in sparse contact regions.
Motivated by Text2Mesh~\cite{michel2022text2mesh}, the semantic alignment loss~$\mathcal{L}_{\mathrm{sem\text{-}align}}$ minimizes the cosine distance between the normalized image and text embeddings.
The segmentation loss $\mathcal{L}_{\mathrm{segm}}$ is the binary cross-entropy loss over the predicted visible head, visible object, and their union.
The image reconstruction loss is $\mathcal{L}_{\mathrm{img}}=\mathcal{L}_{\mathrm{img\text{-}h}}+\mathcal{L}_{\mathrm{img\text{-}ho}}$, where $\mathcal{L}_{\mathrm{img\text{-}h}}$ and $\mathcal{L}_{\mathrm{img\text{-}ho}}$ are L1 distance on the reconstructed hairless head and hairless head--object images, respectively.
The mask prediction loss is $\mathcal{L}_{\mathrm{mask}}=\mathcal{L}_{\mathrm{mask\text{-}h}}+\mathcal{L}_{\mathrm{mask\text{-}o}}$, where both terms are binary cross-entropy losses on the predicted amodal head and object masks.
We set loss weights as $\lambda_{\mathrm{contact}}=1.0$, $\lambda_{\mathrm{sem\text{-}align}}=0.1$, $\lambda_{\mathrm{segm}}=0.2$, $\lambda_{\mathrm{img}}=0.5$, and $\lambda_{\mathrm{mask}}=0.15$.

\input{fig/cohe_contact_dist}

%% file: fig/heco_overall_pipeline.tex
\begin{figure*}[t]
\begin{center}
\includegraphics[width=1.0\linewidth]{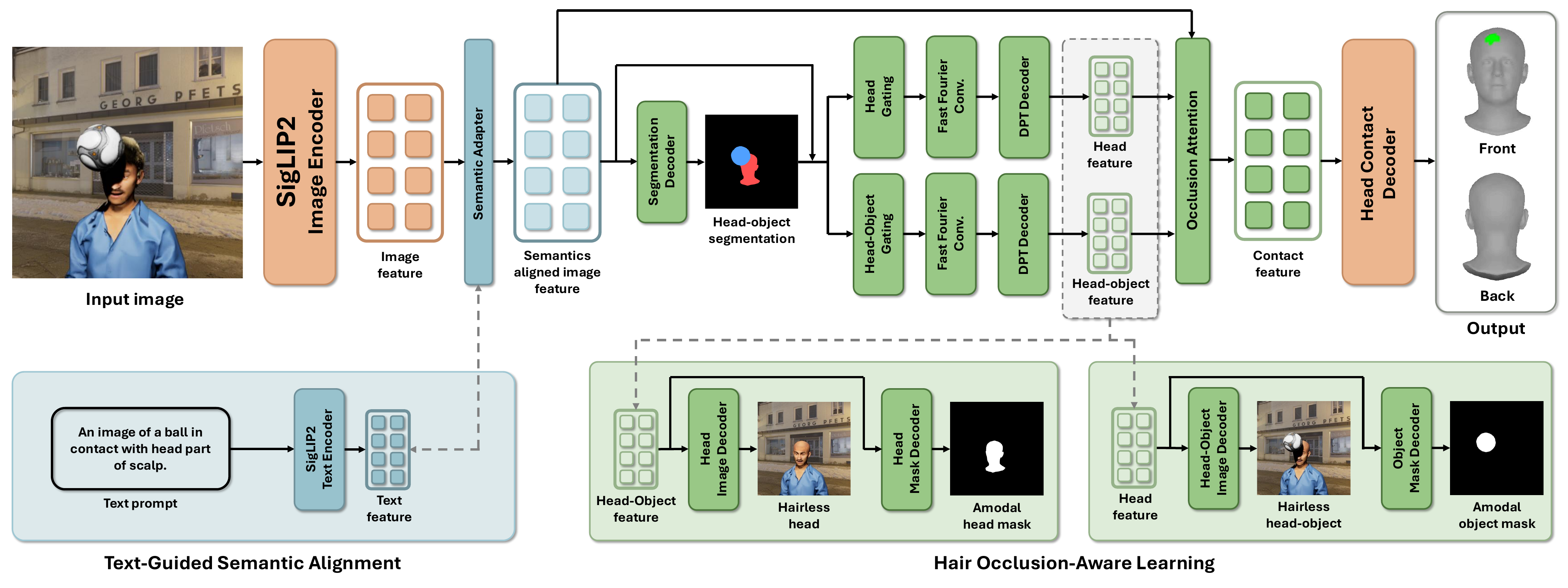}
\end{center}
\vspace{-4mm}
\caption{
\textbf{Overall pipeline of HECO.} Our method encodes image~$\mathbf{I}$ and text prompt~$\mathbf{T}$ with SigLIP2 encoders to build image feature~$\mathbf{F}_{\mathrm{i}}$ and text feature~$\mathbf{F}_{\mathrm{t}}$. After semantic alignment, semantics aligned image feature~$\mathbf{F}_{\mathrm{i}}^{\mathrm{sem}}$ is used for head-object segmentation that gates the aligned image feature~$\mathbf{F}_{\mathrm{i}}^{\mathrm{sem}}$ to build head feature~$\mathbf{F}_{\mathrm{h}}$ and head-object feature~$\mathbf{F}_{\mathrm{ho}}$ which learn hair occlusion through hairless image reconstruction and amodal mask prediction. The head feature~$\mathbf{F}_{\mathrm{h}}$, head-object feature~$\mathbf{F}_{\mathrm{ho}}$, and the aligned image feature~$\mathbf{F}_{\mathrm{i}}^{\mathrm{sem}}$ passes occlusion attention to build contact feature~$\mathbf{F}_{\mathrm{c}}$ which is used by head contact decoder to predict final dense head contact~$\mathbf{C}$.}
\label{fig:pipeline}
\vspace{-0.3cm}
\end{figure*}

%% file: fig/cohe_contact_dist.tex
\begin{figure}[t]
\begin{center}
\includegraphics[width=1.0\linewidth]{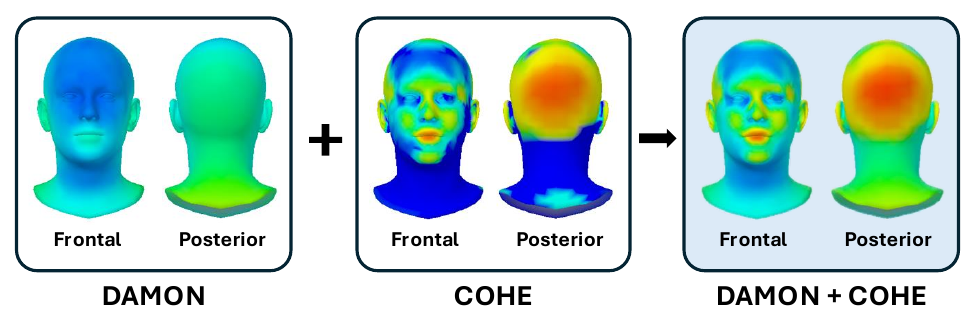}
\end{center}
\vspace{-4mm}
\caption{
\textbf{Contact distribution of COHE.} Our COHE dataset provides additional contact coverage across head regions, such as nose and eyes, that are rarely covered in the existing whole-body contact dataset~\cite{tripathi2023deco}.}
\label{fig:cohe_contact_dist}
\vspace{-0.3cm}
\end{figure}

%% file: sec/4_dataset.tex
\section{Dataset: COHE}

To enable dense head--object contact estimation, we build COHE, a synthetic dataset comprising three complementary data types: capture-based, physics-based, and cross-source interactions.
For capture-based data, we build on the accurate human and object trajectories provided by motion-capture interaction datasets, including GRAB~\cite{taheri2020grab}, OMOMO~\cite{li2023object}, CHAIRS~\cite{jiang2023full}, HODome~\cite{zhang2023neuraldome}, and HIMO~\cite{lv2024himo}.
Since most of these datasets do not provide texture, we apply the human-object appearance transfer and rendering technique of ProciGen~\cite{xie2024template}, while retaining the motion-capture interaction geometry.
This data type therefore preserves realistic interaction trajectories and contact configurations while producing textured images.
For physics-based data, we combine textured BEDLAM2 humans~\cite{tesch2026bedlam2}, MMHead facial expressions~\cite{wu2024mmhead}, and textured object assets from Objaverse~\cite{deitke2023objaverse}, GSO~\cite{downs2022google}, HOPE~\cite{tyree20226}, and Toys4K~\cite{stojanov2021using}.
We only utilize watertight object meshes to support reliable collision and containment analysis as well as contact annotation.
To build data, we randomly sample object instances, trajectories, and target head regions, and use NVIDIA Warp~\cite{Macklin_Warp_A_High-performance_2022} to simulate gravity, collision response, and sustained contact given the human motion geometry.
This pipeline covers diverse interactions such as ball heading, resting on a pillow, face grooming and wiping, mask or helmet wearing, and object impacts.
In Figure~\ref{fig:cohe_contact_dist}, we can observe the contact coverage improvement of head contact.
For cross-source data, we combine the realistic motion of capture-based interactions with the broader human and object asset diversity of the physics-based pipeline.
For human diversity, we transfer a captured head-relative object trajectory to a human motion geometry from physics-based data.
For object replacement, we measure persistent hand--object and head--object functional patches from the source interaction and describe each interaction candidate surface using its spatial spectrum, normal variation, scale-normalized principal curvature, and patch size, to align the most compatible replacement patch to the measured source patch.
The aligned object is transported in the supporting-hand frame and physically simulated under gravity, bounded grasp constraints, and head/hand collision response, rather than simply replaying the source object root motion.
We retain a generated sequence only when its functional contact, grasp stability, temporal contact pattern, and penetration checks remain valid throughout the interaction.
Unlike ProciGen~\cite{xie2024template}, which transfers interactions through category-specific shape correspondences, our curvature-driven replacement does not require the source and target objects to share a category and can therefore generate interactions with compatible objects from novel categories.
Across all three data types, head contact is represented on the FLAME topology~\cite{li2017learning} and is annotated from the posed head and object geometry using a 5-mm surface-distance threshold, with enclosed head vertices additionally treated as contact.
Overall, COHE contains 18,632 training samples and 1,668 test samples where the training set consists of 8,118 physics-based, 9,164 capture-based, and 1,350 cross-source samples.

%% file: sec/5_implementation_details.tex
\section{Implementation details}
PyTorch~\cite{paszkepytorch} is used for implementation.
Our image backbone is initialized from the publicly released SigLIP2 Giant model~\cite{tschannen2025siglip}, pretrained on WebLI image--text data~\cite{chen2022pali}, and processes images at a resolution of $384\times384$.
During training, we apply PASTA amplitude-spectrum augmentation~\cite{chattopadhyay2023pasta} to synthetic input images with $\alpha=3.0$, $\beta=0.25$, and $k=2$.
We use the AdamW optimizer~\cite{loshchilov2018decoupled} with learning rates of $10^{-5}$ for the SigLIP2 image encoder and $10^{-4}$ for the remaining trainable modules, and a weight decay of $10^{-4}$ and a mini-batch size of $8$.
The learning rate is decayed by a factor of $0.9$ after the 5th epoch.
We train HECO for $10$ epochs on a single NVIDIA RTX A6000 GPU.

%% file: sec/6_experiments.tex
\section{Experiments}

\subsection{Datasets}
We use our proposed COHE dataset and DAMON~\cite{tripathi2023deco} for training and evaluation.
During training, we sample COHE and DAMON at a $1{:}1$ ratio.
DAMON provides real-world human--object interaction images with dense contact annotations defined on the SMPL-X~\cite{pavlakos2019expressive} human mesh.
We convert the DAMON annotations to FLAME~\cite{li2017learning} using the official SMPL-X to FLAME vertex correspondence.
DAMON is used only for dense head contact supervision, as the other auxiliary annotations are not available.
For DAMON evaluation, we exclude one of the 785 official test samples because the target person has no visible head region.
Among the remaining 784 samples, we compute precision, recall, and F1 on the 129 samples that contain at least one head-contact vertex.

\input{fig/qualitative_dense_head_contact}

\subsection{Evaluation metrics}
To evaluate dense head--object contact estimation, we report precision, recall, and F1 score on the $V=5{,}023$ vertices of the FLAME head mesh.
We use the full-resolution contact probabilities $\hat{\mathbf{C}}_1=\sigma(\mathbf{C}_1)\in[0,1]^V$, corresponding to the finest level ($i=1$) of the multi-level outputs, and binary ground truth $\mathbf{C}_{\mathrm{GT}}\in\{0,1\}^V$.
We threshold $\hat{\mathbf{C}}_1$ at $0.5$ and compute precision and recall for each frame.
We report their averages over samples containing at least one ground-truth contact vertex with the reported F1 score computed as their harmonic mean.

\subsection{Ablation study}

\noindent\textbf{Effectiveness of COHE dataset.}
Table~\ref{tab:abl_train_data} compares the effects of physics-based, capture-based, and cross-source training data in the COHE dataset.
Physics-based and capture-based data alone achieve F1-scores of 0.327 and 0.092, respectively.
Combining physics-based and capture-based data achieves an F1-score of 0.386, suggesting complementary coverage of head-object interactions.
Including cross-source data yields the highest F1-score of 0.429, improving the score by 11.14\% over the configuration without cross-source data.
In particular, the results show that relying solely on capture-based data leads to poor performance due to the gap in contact distributions between capture-based and physics-based data, highlighting the usefulness of physics-based data.
Cross-source data further provides valuable head--object contact samples involving novel object types.
Table~\ref{tab:abl_domain_gen} evaluates the inclusion of COHE dataset with DAMON~\cite{tripathi2023deco} dataset.
Compared with DAMON-only training, joint training increases F1-score from 0.434 to 0.481.
The higher F1-score suggests that COHE can complement real-image supervision by improving contact coverage.
In the end, COHE serves as a valuable dataset for head-object interaction estimation.

\input{tab/abl_train_data}

\input{tab/abl_domain_gen}

\noindent\textbf{Effectiveness of text-guided semantic alignment.}
Table~\ref{tab:abl_align} evaluates text-guided semantic alignment on COHE dataset.
Compared with HECO without semantic alignment, the full HECO model increases F1-score from 0.320 to 0.429.
This shows that semantic guidance effectively improves head-object contact estimation by encouraging image features to retain contact-relevant semantic information that supports more accurate and generalizable head contact estimation.
Additionally, Table~\ref{tab:sota_semantic_feat} compares CLIP~\cite{radford2021learning}, BLIP~\cite{li2022blip}, EVA-CLIP~\cite{sun2023eva}, SigLIP~\cite{zhai2023sigmoid}, PE-Core~\cite{bolya2026perception}, and SigLIP2~\cite{tschannen2025siglip}, replacing both image and text encoders.
SigLIP2 achieves the highest reported F1-score of 0.429, outperforming all other methods.
This supports our use of SigLIP2 as encoders for semantic alignment.

\input{tab/abl_alignment}

\input{tab/sota_semantic_feat}

\noindent\textbf{Effectiveness of occlusion-aware contact learning.}
Table~\ref{tab:abl_inpaint} examines two-stream architecture that builds head feature and head-object feature, amodal mask prediction that predicts amodal head mask and amodal object mask, and hairless image reconstruction that reconstructs hairless head image and hairless head-object image.
Introducing two streams increases F1-score from 0.287 to 0.300, and adding amodal-mask prediction further increases it to 0.327. 
Full configuration with hairless image reconstruction achieves the highest F1-score of 0.429.
The improvements suggest that separating head and head--object features and learning occlusion-aware targets in images and masks provide useful evidence for learning dense head contact.

\input{tab/abl_inpaint}

\noindent\textbf{Effectiveness of occlusion attention.}
Table~\ref{tab:abl_attn} compares occlusion attention with simple feature fusion.
The attention-based configuration increases F1-score from 0.380 to 0.429. 
The large increase in recall shows that adaptive integration of head, head--object, and semantic-aligned image features helps recover contact regions missed by simple fusion.

\input{tab/abl_attention}

\subsection{Comparison with state-of-the-art methods}

\noindent\textbf{Dense head contact estimation.}
Table~\ref{tab:sota_head_cont} compares HECO with POSA~\cite{hassan2021populating}, BSTRO~\cite{huang2022capturing}, DECO~\cite{tripathi2023deco}, and InteractVLM~\cite{dwivedi2025interactvlm} on DAMON~\cite{tripathi2023deco}.
HECO achieves the highest precision, recall, and F1-score of 0.381, 0.653, and 0.481, respectively.
POSA, BSTRO, and DECO yield low head-contact F1-scores of 0.051, 0.000, and 0.050 due to their almost non-existence or highly inaccurate prediction of head contact.
InteractVLM achieves an F1-score of 0.093 using LLaVA~\cite{liu2023visual}-predicted object categories (LLaVA obj) and 0.262 using ground-truth object categories (GT obj).
Note that using ground-truth object categories is the default setting of InteractVLM.
HECO outperforms InteractVLM with GT object categories in F1-score by 83.59\% increase without requiring GT object category text during inference.
This shows that HECO can serve as an accurate head contact estimation method replacing previous methods.

\input{tab/sota_head_cont}

%% file: fig/qualitative_dense_head_contact.tex
\begin{figure*}[t]
\begin{center}
\includegraphics[width=1.0\linewidth]{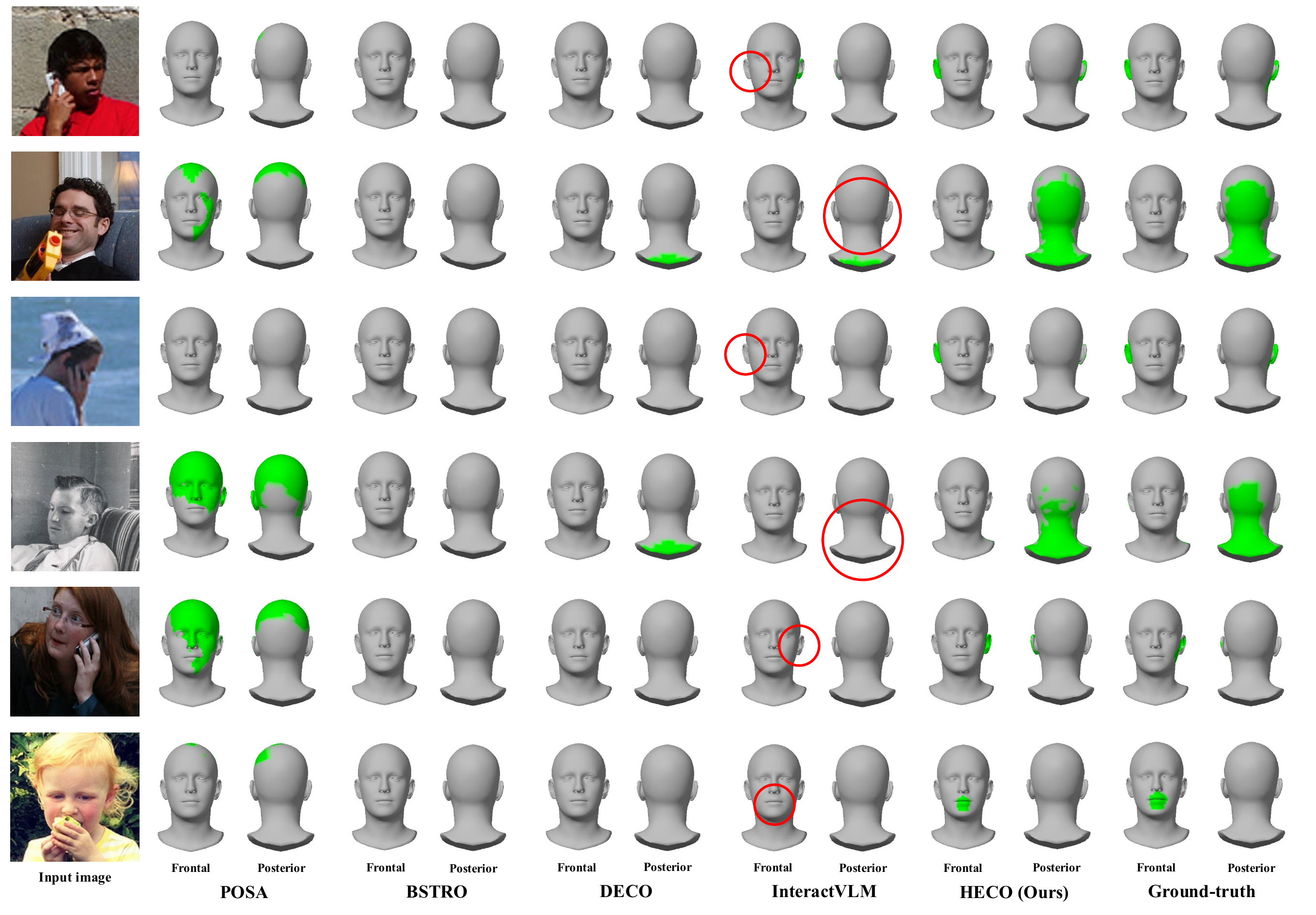}
\end{center}
\vspace{-4mm}
\caption{
\textbf{Qualitative comparison of dense head contact estimation with POSA~\cite{hassan2021populating}, BSTRO~\cite{huang2022capturing}, DECO~\cite{tripathi2023deco}, InteractVLM~\cite{dwivedi2025interactvlm} on DAMON~\cite{tripathi2023deco} dataset.} Red circles indicate exemplar regions that HECO outperforms previous methods.}
\label{fig:sota_qual_dense_head}
\vspace{-0.3cm}
\end{figure*}

%% file: tab/abl_train_data.tex
\begin{table}[hbtp]
\centering
\footnotesize
\setlength{\tabcolsep}{2.7pt}
\caption{\textbf{Ablation on data type of COHE dataset.}}
\vspace{-0.5em}
\begin{tabular}{@{}cccccc@{}}
\toprule
Capture & Physics & Cross-source & Precision $\uparrow$ & Recall $\uparrow$ & F1-Score $\uparrow$ \\
\midrule
\ding{51} & \ding{55} & \ding{55} & 0.065 & 0.156 & 0.092 \\
\ding{55} & \ding{51} & \ding{55} & 0.270 & 0.414 & 0.327 \\
\ding{51} & \ding{51} & \ding{55} & 0.328 & 0.468 & 0.386 \\
\ding{51} & \ding{51} & \ding{51} & \textbf{0.377} & \textbf{0.498} & \textbf{0.429} \\
\bottomrule
\end{tabular}
\label{tab:abl_train_data}
\vspace{-0.2cm}
\end{table}

%% file: tab/abl_domain_gen.tex
\begin{table}[hbtp]
\centering
\footnotesize
\setlength{\tabcolsep}{3.5pt}
\caption{\textbf{Ablation of training dataset on DAMON~\cite{tripathi2023deco} dataset.}}
\vspace{-0.5em}
\begin{tabular}{@{}ccccc@{}}
\toprule
DAMON & COHE dataset & Precision $\uparrow$ & Recall $\uparrow$ & F1-Score $\uparrow$ \\ 
\midrule
\ding{51} & \ding{55} & \textbf{0.422} & 0.447 & 0.434 \\
\ding{51} & \ding{51} & 0.381 & \textbf{0.653} & \textbf{0.481} \\
\bottomrule
\end{tabular}
\label{tab:abl_domain_gen}
\vspace{-0.2cm}
\end{table}

%% file: tab/abl_alignment.tex
\begin{table}[hbtp]
\centering
\footnotesize
\setlength{\tabcolsep}{2.5pt}
\renewcommand{\arraystretch}{0.9}
\caption{\textbf{Ablation on text-guided semantic alignment on COHE dataset.}}
\vspace{-0.5em}
\begin{tabular}{@{}lccc@{}}
\toprule
Methods & Precision $\uparrow$ & Recall $\uparrow$ & F1-Score $\uparrow$ \\
\midrule
w/o Semantic alignment & 0.288 & 0.361 & 0.320 \\
w/ Semantic alignment~(Ours) & \textbf{0.377} & \textbf{0.498} & \textbf{0.429} \\
\bottomrule
\end{tabular}
\label{tab:abl_align}
\vspace{-0.2cm}
\end{table}

%% file: tab/sota_semantic_feat.tex
\begin{table}[hbtp]
\centering
\footnotesize
\setlength{\tabcolsep}{4pt}
\caption{\textbf{Comparison of semantic alignment on COHE dataset.}}
\vspace{-0.5em}
\begin{tabular}{@{}lccc@{}}
\toprule
Method & Precision $\uparrow$ & Recall $\uparrow$ & F1-Score $\uparrow$ \\ 
\midrule
CLIP~\cite{radford2021learning} & 0.278 & 0.329 & 0.301 \\ 
BLIP~\cite{li2022blip} & 0.250 & 0.371 & 0.299 \\ 
EVA-CLIP~\cite{sun2023eva} & 0.337 & 0.433 & 0.379  \\ 
SigLIP~\cite{zhai2023sigmoid} & 0.375 & 0.425 & 0.398 \\ 
PE-Core~\cite{bolya2026perception} & 0.302 & 0.405 & 0.346 \\ 
SigLIP2~\cite{tschannen2025siglip} & \textbf{0.377} & \textbf{0.498} & \textbf{0.429} \\
\bottomrule
\end{tabular}
\label{tab:sota_semantic_feat}
\vspace{-0.2cm}
\end{table}

%% file: tab/abl_inpaint.tex
\begin{table}[hbtp]
\centering
\footnotesize
\setlength{\tabcolsep}{2.5pt}
\renewcommand{\arraystretch}{0.9}
\caption{\textbf{Ablation on hair occlusion-aware learning on COHE dataset.}}
\vspace{-0.5em}
\resizebox{\columnwidth}{!}{%
\begin{tabular}{@{}cccccc@{}}
\toprule
Two stream & Amodal mask & Hairless image & Precision $\uparrow$ & Recall $\uparrow$ & F1-Score $\uparrow$ \\
\midrule
\ding{55} & \ding{55} & \ding{55} & 0.282 & 0.291 & 0.287 \\
\ding{51} & \ding{55} & \ding{55} & 0.278 & 0.326 & 0.300 \\
\ding{51} & \ding{51} & \ding{55} & 0.300 & 0.358 & 0.327 \\
\ding{51} & \ding{51} & \ding{51} & \textbf{0.377} & \textbf{0.498} & \textbf{0.429} \\
\bottomrule
\end{tabular}
}
\label{tab:abl_inpaint}
\vspace{-0.2cm}
\end{table}

%% file: tab/abl_attention.tex
\begin{table}[hbtp]
\centering
\footnotesize
\setlength{\tabcolsep}{2.5pt}
\renewcommand{\arraystretch}{0.9}
\caption{\textbf{Ablation on occlusion attention on COHE dataset.}}
\vspace{-0.5em}
\begin{tabular}{@{}lccc@{}}
\toprule
Methods & Precision $\uparrow$ & Recall $\uparrow$ & F1-Score $\uparrow$ \\
\midrule
w/o Occlusion attention & 0.366 & 0.396 & 0.380 \\
w/ Occlusion attention (Ours) & \textbf{0.377} & \textbf{0.498} & \textbf{0.429} \\
\bottomrule
\end{tabular}
\label{tab:abl_attn}
\vspace{-0.2cm}
\end{table}

%% file: tab/sota_head_cont.tex
\begin{table}[hbtp]
\centering
\footnotesize
\setlength{\tabcolsep}{2.8pt}
\renewcommand{\arraystretch}{0.9}
\caption{\textbf{Comparison with SOTA methods of dense head contact estimation on DAMON~\cite{tripathi2023deco} dataset.} LLaVA obj refers to object type predicted by LLaVA and GT obj refers to GT object type. The default setting of InteractVLM is GT obj.}
\vspace{-0.5em}
\begin{tabular}{@{}lccc@{}}
\toprule
Methods & Precision $\uparrow$ & Recall $\uparrow$ & F1-Score $\uparrow$ \\ 
\midrule
POSA~\cite{hassan2021populating} & 0.040 & 0.068 & 0.051 \\
BSTRO~\cite{huang2022capturing} & 0.000 & 0.000 & 0.000 \\
DECO~\cite{tripathi2023deco} & 0.061 & 0.042 & 0.050 \\
InteractVLM~\cite{dwivedi2025interactvlm} (LLaVA~\cite{liu2023visual} obj) & 0.132 & 0.072 & 0.093 \\
InteractVLM~\cite{dwivedi2025interactvlm} (GT obj) & 0.318 & 0.223 & 0.262 \\
\midrule
HECO~(Ours) & \textbf{0.381} & \textbf{0.653} & \textbf{0.481} \\
\bottomrule
\end{tabular}
\label{tab:sota_head_cont}
\vspace{-0.2cm}
\end{table}

%% file: sec/7_conclusion.tex
\section{Conclusion}
We propose HECO, a powerful framework for dense head--object contact estimation that learns from synthetic data while addressing diverse interactions and hair occlusion.
To capture semantic contact, we incorporate semantic context from VLM text encoder that contains head contact semantics.
To discover contact hidden by hair, we develop hair occlusion-aware learning via hair removal and amodal mask prediction.
Together with our proposed COHE dataset, HECO outperforms previous state-of-the-art methods by a significant margin, demonstrating the potential of synthetic data for understanding head--object contact.

%% file: sec/X_suppl.tex
\clearpage
\maketitlesupplementary

\appendix

\setcounter{section}{0}
\setcounter{table}{0}
\setcounter{figure}{0}

\renewcommand{\thesection}{S\arabic{section}}
\renewcommand{\thetable}{S\arabic{table}}   
\renewcommand{\thefigure}{S\arabic{figure}}

\renewcommand{\theHsection}{S\arabic{section}}
\renewcommand{\theHtable}{S\arabic{table}}
\renewcommand{\theHfigure}{S\arabic{figure}}

In this supplementary material, we present further technical details and experimental results that could not be included in the main manuscript due to space limitations.
The supplementary material is organized as follows:
\begin{itemize}
\item \ref{sec:supp_cohe_config}. Configuration of COHE dataset
\item \ref{sec:supp_data_gen}. Details of dataset generation
\item \ref{sec:supp_dense_head_label}. Details of dense head contact labels
\item \ref{sec:supp_vis_hair_occ}. Visualization of hair occlusion-aware learning
\item \ref{sec:more_sota_comp}. More comparisons with state-of-the-art methods
\item \ref{sec:supp_quant_backbone}. Quantitative results on different backbones
\item \ref{sec:supp_comp_req}. Computational requirements
\item \ref{sec:supp_more_qual}. More qualitative results
\item \ref{sec:supp_limit}. Limitations and societal impacts
\end{itemize}

\section{Configuration of COHE dataset}
\label{sec:supp_cohe_config}
The proposed COHE dataset is a synthetic dataset for dense head--object contact estimation, providing RGB images paired with ground-truth dense contact annotations on the FLAME~\cite{li2017learning} head model mesh.
The dataset covers diverse head contact regions, including the scalp, forehead, eyes, ears, nose, cheeks, and mouth, as well as non-contact scenarios.
As illustrated in Figure~\ref{fig:supp_cohe_contact_gt}, COHE captures a wide range of head--object interactions, from localized contact around small facial regions to broad contact across the head surface.
For instance, interactions with pillows involve extensive scalp contact, whereas eyeglasses, headphones, and mug exhibit localized contact around the nose, ears, and mouth, respectively.
These examples demonstrate the diversity of head region specific contact patterns represented in the dataset.
To capture diverse head--object interactions, COHE contains 46 object categories organized into seven groups, including drinkware, home \& furniture, personal care, sports \& leisure, wearables \& accessories, devices \& equipment, and food \& produce.
As shown in Figure~\ref{fig:supp_cohe_obj_cat}, drinkware and home and furniture account for the largest proportions of training samples, followed by personal care.
The dataset includes frequently encountered objects such as pillows, eyeglasses, mugs, cups, and electric shavers, along with less frequent object categories, providing diverse head interaction scenarios.
Furthermore, different object groups exhibit distinct contact patterns across head regions.
As shown in Figure~\ref{fig:supp_cohe_obj_rel}, home \& furniture objects are predominantly associated with scalp contact, whereas drinkware primarily involves mouth \& chin contact.
Overall, mouth \& chin contact accounts for the largest proportion, followed by scalp and ears.
These distributions demonstrate that COHE captures diverse semantic relationships between interacting objects and head contact regions, providing valuable supervision for dense head--object contact estimation.

\section{Details of dataset generation}
\label{sec:supp_data_gen}

Human--object interactions in COHE are rendered in Unreal Engine 5~\cite{unrealengine5} with variations in human appearance, object materials, HDR illumination, and camera settings.
Camera placement is checked using projected head, contact, and object vertices to limit truncation and excessively small interactions in the image.

Capture-based data first prepares human scan textures appearance to be baked onto the SMPL-X UV layout so that the transferred appearance follows the recorded body animation.
When scan-derived human surface offsets for appearance are used, they are transferred to the SMPL-X mesh through barycentric correspondence and rotated with the body joints at each frame.
Object appearance is prepared by baking donor textures onto the original interaction mesh as in ProciGen~\cite{xie2024template}, or by using an existing texture atlas, with roughness, metallic, and available normal or opacity maps assigned in the renderer.
This separates object material assignment from geometry replacement, allowing appearance changes without altering the object surface used for contact annotation.

\input{fig/supp_cohe_contact_gt}

\input{fig/supp_cohe_object_categories}

\input{fig/supp_cohe_object_relationships}

Physics-based data generates object trajectory by using interaction-specific target locations and motion parameters.
To preserve appearance alignment in human, clothes, and hair, BEDLAM2~\cite{tesch2026bedlam2} humans use matching body, clothing, and hair-binding assets with synchronized animation frames and coordinate transformations.
Ball heading and rigid impacts begin with a target head surface location, contact time, and incoming speed and direction, from which the initial object position and velocity are computed under gravity in NVIDIA Warp~\cite{Macklin_Warp_A_High-performance_2022}.
Subsequent motion follows collision response against the posed human meshes, with mass, friction, and restitution controlling the response.
Grooming trajectories follow short curved strokes whose direction and length are determined within a selected facial or scalp region.
The tool's working surface is oriented toward the local skin normal, and mesh collision queries adjust its distance from the skin during approach, contact, and withdrawal.
Wiping trajectories use surface waypoints within the selected facial region to guide the cloth attachment, while cloth constraints and collision handling determine the remaining fabric motion.
Resting interactions instead begin with a fitted support placement beneath the head, checked against the body meshes.

Cross-source data generation first expresses captured object poses relative to the source head and maps them into the recipient head coordinate system.
Object replacement aligns the replacement's functional surface to the source contact patch, which we call a curvature-driven object replacement.
Supporting-hand motion then supplies attachment targets for the physical simulation.

To provide supervision for hair occlusion-aware learning, we generate a hairless human-only image and a hairless human--object image.
Both preserve the input scene's pose, camera, illumination, and background, with the object removed for the former and retained for the latter.
In the physics-based data type, the BEDLAM2 hair grooms are disabled during these passes.
In the capture-based data type, hair embedded in the scalp texture or mesh are directly modified by the scalp receiving a bald texture matched to facial skin color, and hair-related scalp deformations removed.
Geometry changes are confined to the scalp and a narrow transition along its boundary, preserving the remaining head and body regions.

Amodal masks are generated by projecting the complete 3D head and object meshes onto the 2D image plane using the rendering camera parameters.
Each mesh is projected separately without hair or other scene objects blocking it, so its mask includes both visible and occluded regions.
Modal masks use the same camera and retain only surfaces visible after scene occlusion is applied.
To check image--geometry alignment, we compare projected human--object silhouettes with the foreground silhouettes in the rendered images.
To generate dense contact annotation, we follow the procedure mentioned in Section~\ref{sec:supp_dense_head_label}.

\input{tab/supp_comp_require}

\section{Details of dense head contact labels}
\label{sec:supp_dense_head_label}

Following previous work on dense hand contact estimation~\cite{jung2025learning}, we obtain dense head contact labels using geometry-based thresholding with the Trimesh library~\cite{trimesh}.
For all three data types of COHE dataset, we use head and interacting object meshes in a common 3D world coordinate system measured in meters.
Dense head contact labels are represented on the FLAME topology with 5,023 vertices.
Since COHE is constructed from human--object interactions represented by SMPL-X full-body meshes, we extract the FLAME mesh topology using a fixed, ordered SMPL-X-to-FLAME vertex correspondence.
For each head vertex, we compute its minimum Euclidean distance to the object's triangular surface and assign a positive label when the distance is at most 5mm.
For watertight object meshes with valid containment queries, vertices inside the object are also labeled positive.
Penetration quality control is applied separately from contact labeling.
The standard head-geometry check tolerates penetration up to 2mm, flags depths greater than 2mm and up to 5mm for review, and treats depths above 5mm as severe penetration.
The final acceptance as COHE dataset is determined after manual review for flagged data and automatic pass for tolerated penetration.
In the end, each sample receives a binary vector that annotates contact across all 5,023 vertices on FLAME head mesh topology.

\section{Visualization of hair occlusion-aware learning}
\label{sec:supp_vis_hair_occ}
Figure~\ref{fig:supp_hair_occ_inpaint_qual} presents qualitative results of our hair occlusion-aware learning, which reconstructs hairless head and hairless head--object images from input images.
The reconstructed images produced by HECO closely resemble the ground truth, demonstrating its ability to infer the underlying head appearance despite hair occlusion.
In particular, the hairless head reconstruction removes both hair and interacting objects, while the hairless head--object reconstruction preserves the interacting objects while removing hair.
The reconstructions generally preserve the underlying head structure and spatial relationships between the head and interacting objects.
Figure~\ref{fig:supp_hair_occ_mask_qual} shows qualitative results of amodal mask prediction.
The predicted amodal masks generally align well with the ground truth, capturing the underlying head regions beyond visible boundaries.
Similarly, the predicted modal masks robustly capture the visible regions.
The predictions demonstrate that HECO effectively learns spatial representations beyond directly observable image regions.
Together, these reconstruction and mask prediction results suggest that hair occlusion-aware learning provides useful representations of the underlying head geometry and head--object spatial relationships, supporting dense head contact estimation under hair occlusion.

\section{More comparisons with state-of-the-art methods}
\label{sec:more_sota_comp}

In the main paper, we compare our method with state-of-the-art (SOTA) methods using their officially released model checkpoints.
However, as these methods were trained using different dataset configurations, the comparison does not isolate differences in model architecture.
Specifically, POSA~\cite{hassan2021populating} was trained on PROX~\cite{hassan2019resolving}, BSTRO~\cite{huang2022capturing} on RICH~\cite{huang2022capturing}, DECO~\cite{tripathi2023deco} on DAMON~\cite{tripathi2023deco}, RICH, and PROX, and InteractVLM~\cite{dwivedi2025interactvlm} on DAMON.
To ensure a fair comparison of model architectures, we train all methods for contact prediction on the same DAMON and COHE training sets using their respective pre-trained initializations.
Table~\ref{tab:supp_sota_body_cont_rep} compares the SOTA methods under the same training dataset setting while preserving their original dense body contact estimation architectures.
Table~\ref{tab:supp_sota_head_cont_rep} further compares the SOTA methods under the same training dataset setting after adapting their dense body contact architectures for dense head contact estimation, which we denote as head version.
For the head version experiments, all methods predict contact on the 5,023 FLAME vertices.
Specifically, BSTRO uses a FLAME hierarchy with 431, 1,723, and 5,023 contact tokens and multiscale contact losses, POSA replaces its SMPL-X graph with a FLAME mesh hierarchy in both the encoder and decoder, DECO replaces its body-part supervision with 22-class FLAME head-region supervision while retaining scene-context supervision, and InteractVLM uses head-region prompts and a multi-view 2D contact mask rendering in FLAME head model to predict contact on the FLAME vertices.
As shown in Table~\ref{tab:supp_sota_body_cont_rep}, HECO achieves the highest F1-score of 0.481, outperforming the strongest baseline, DECO, by 97.94\%.
Similarly, Table~\ref{tab:supp_sota_head_cont_rep} shows that HECO surpasses the strongest baseline, BSTRO, by 68.77\% in F1-score.
These results demonstrate that HECO consistently outperforms previous methods even when trained on the same datasets and adapted for dense head contact estimation.

\section{Quantitative results on different backbones}
\label{sec:supp_quant_backbone}
We evaluate the performance of HECO using various SigLIP2~\cite{tschannen2025siglip} backbone variants on the DAMON~\cite{tripathi2023deco} dataset, keeping all other components fixed.
In Table~\ref{tab:supp_diff_back}, SigLIP2-Giant achieves the highest precision, recall, and F1-score, demonstrating the strongest overall performance among all backbones.
SigLIP2-So400m achieves the second highest F1-score, followed by SigLIP2-Large and SigLIP2-Base.
These results indicate that larger backbone models generally improve dense head contact estimation, particularly in precision.
Notably, recall remains relatively consistent across backbone variants, while precision shows a more substantial improvement with increasing model capacity.
This suggests that larger backbones are more effective at distinguishing contact from non-contact regions, rather than simply increasing contact coverage.
Overall, these results demonstrate the benefit of larger vision-language backbones for dense head contact estimation and justify the selection of SigLIP2-Giant as the default backbone in our HECO model.

\input{tab/supp_sota_body_ver_cont_rep}
\input{tab/supp_sota_head_ver_cont_rep}

\section{Computational requirements}
\label{sec:supp_comp_req}
Table~\ref{tab:supp_comp_require} reports the computational requirements of HECO under different SigLIP2~\cite{tschannen2025siglip} backbone configurations.
SigLIP2-Giant delivers the strongest dense head contact estimation performance, as shown in Table~\ref{tab:supp_diff_back}, but incurs the highest computational overhead, requiring approximately 27.6 GB of training memory and 12.9 GB of inference memory.
SigLIP2-So400m and SigLIP2-Large provide more moderate alternatives, reducing memory consumption and computational cost while maintaining competitive estimation performance.
Smaller variant also achieves faster inference, highlighting the trade-off between model capacity and computational efficiency.
Among all configurations, SigLIP2-Base offers the most efficient configuration, achieving the fastest inference speed of 5.08 fps with substantially fewer parameters and lower memory requirements.
Overall, these results demonstrate the flexibility of HECO across different backbone configurations, allowing computational efficiency to be balanced against estimation accuracy.

\input{tab/supp_diff_back}

\section{More qualitative results}
\label{sec:supp_more_qual}
Figure~\ref{fig:supp_more_qual1} and Figure~\ref{fig:supp_more_qual2} present additional qualitative comparisons with POSA~\cite{hassan2021populating}, BSTRO~\cite{huang2022capturing}, DECO~\cite{tripathi2023deco}, and InteractVLM~\cite{dwivedi2025interactvlm} on the DAMON~\cite{tripathi2023deco} dataset.
Our HECO consistently produces more accurate contact predictions across diverse head--object interactions compared with previous methods.
POSA frequently predicts excessive contact regions, particularly around the scalp and forehead, even when the actual contact occurs in localized facial regions.
For example, in the first three rows of Figure~\ref{fig:supp_more_qual1}, POSA predicts contact around the scalp or fails to localize contact, whereas the ground truth indicates localized contact around the mouth.
Similarly, in the fourth, sixth, and eighth rows of Figure~\ref{fig:supp_more_qual2}, POSA predicts substantially broader contact regions than the ground truth.
In contrast, BSTRO and DECO frequently fail to detect head contact, producing nearly empty contact predictions across most samples.
InteractVLM detects contact in some cases but often predicts inaccurate contact locations, particularly around the neck and lower face.
For instance, in the fourth row of Figure~\ref{fig:supp_more_qual1}, InteractVLM predicts contact across the neck despite the ground truth indicating contact primarily on the posterior head.
Furthermore, in the sixth row of Figure~\ref{fig:supp_more_qual2}, InteractVLM fails to detect localized ear contact despite its presence in the ground truth.
In contrast, HECO accurately localizes contact across diverse head regions, including the lips, cheeks, ears, and posterior head, while reducing excessive contact predictions.
These results highlight the effectiveness of our text-guided semantic alignment and hair occlusion-aware learning in capturing diverse head contact patterns.

\section{Limitations and societal impacts}
\label{sec:supp_limit}
\noindent\textbf{Limitations.}
Our HECO addresses two key challenges in dense head--object contact estimation: diverse contact patterns across head regions and hair occlusion.
Furthermore, we introduce COHE, a synthetic dataset that provides diverse head--object interactions with dense contact annotations.
Nevertheless, several limitations remain.
First, HECO primarily operates on cropped head images, which may exclude useful contextual information from the full image, particularly when the head or interacting object is heavily occluded.
Incorporating full-image context or integrating HECO with dense full-body contact estimation methods could improve robustness under such conditions.
Second, HECO estimates contact from a single RGB image without explicitly modeling temporal information.
Temporal cues, such as object motion and changes in contact states, could help distinguish actual contact from occlusion.
Although COHE contains temporal interaction sequences, HECO does not yet exploit their temporal relationship.
Extending HECO to video-based dense head contact estimation is therefore a promising direction for future research.
Third, HECO focuses solely on estimating contact regions on the head surface, without explicitly predicting the corresponding contact regions on interacting objects.
Extending HECO to jointly estimate dense contact on both the head and object surfaces could provide a more comprehensive understanding of head--object interactions and their spatial relationships.
Lastly, although COHE provides diverse head--object interactions, its synthetic appearances do not yet fully capture the visual realism of real-world images.
Employing more advanced graphics engines and higher-quality 3D assets could improve the photorealism of COHE, further reducing the synthetic-to-real domain gap and potentially improving HECO's generalization and estimation accuracy.

\noindent\textbf{Societal impacts.}
Our HECO has potential applications in sports analytics, human safety, AR/VR, and human--object interaction understanding.
In particular, dense head contact estimation could support the analysis of head impacts and interactions involving protective equipment.
However, deploying such systems in real-world environments raises concerns regarding privacy, safety, and potential misuse.
Since head contact estimation relies on images containing human faces, data collection and processing should respect individual privacy through informed consent, limited data retention, and appropriate safeguards.
Furthermore, inaccurate contact predictions could lead to incorrect interpretations of head impacts, particularly in safety-critical applications.
HECO is not designed for medical diagnosis or injury assessment, and such applications would require rigorous validation, calibrated uncertainty estimates, and expert oversight.

\clearpage

\input{fig/supp_hair_occ_amodal_inpaint_qual}

\clearpage

\input{fig/supp_hair_occ_amodal_mask_qual}

\clearpage

\input{fig/supp_more_qualitative1}

\clearpage

\input{fig/supp_more_qualitative2}

%% file: fig/supp_cohe_contact_gt.tex
\begin{figure*}[t]
\begin{center}
\includegraphics[width=1.0\linewidth]{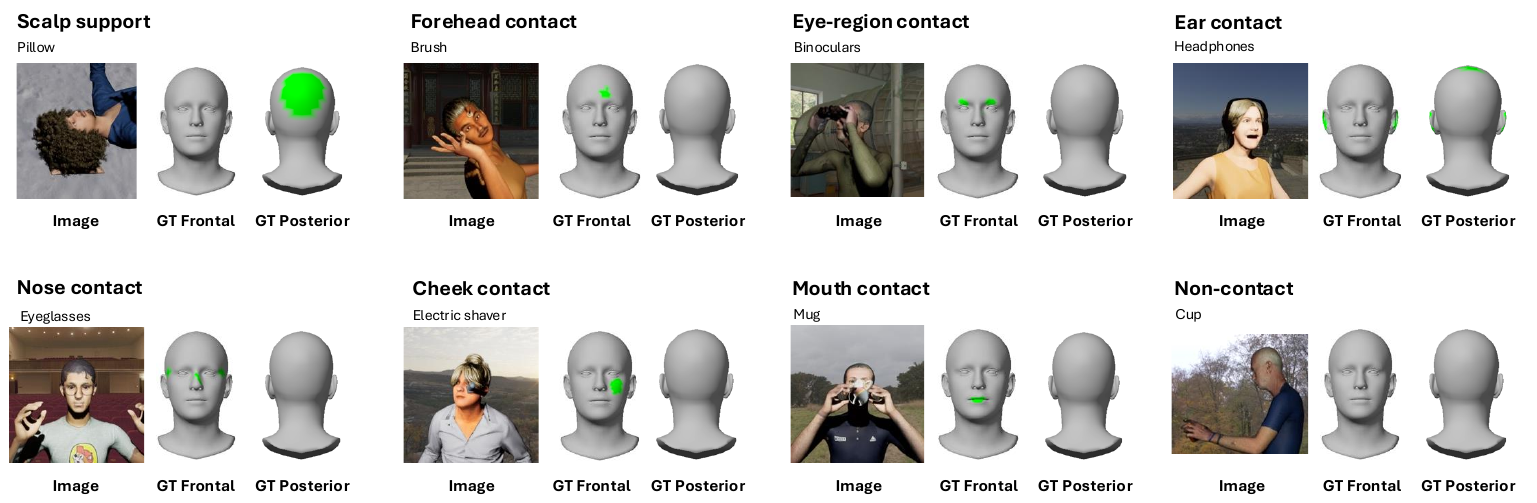}
\end{center}
\vspace{-4mm}
\caption{
\textbf{Ground-truth dense head contact annotation in COHE Dataset.} Our COHE dataset provides RGB image along with its paired dense head contact annotation in FLAME head model mesh topology.}
\label{fig:supp_cohe_contact_gt}
\end{figure*}

%% file: fig/supp_cohe_object_categories.tex
\begin{figure}[t]
\begin{center}
\includegraphics[width=1.0\linewidth]{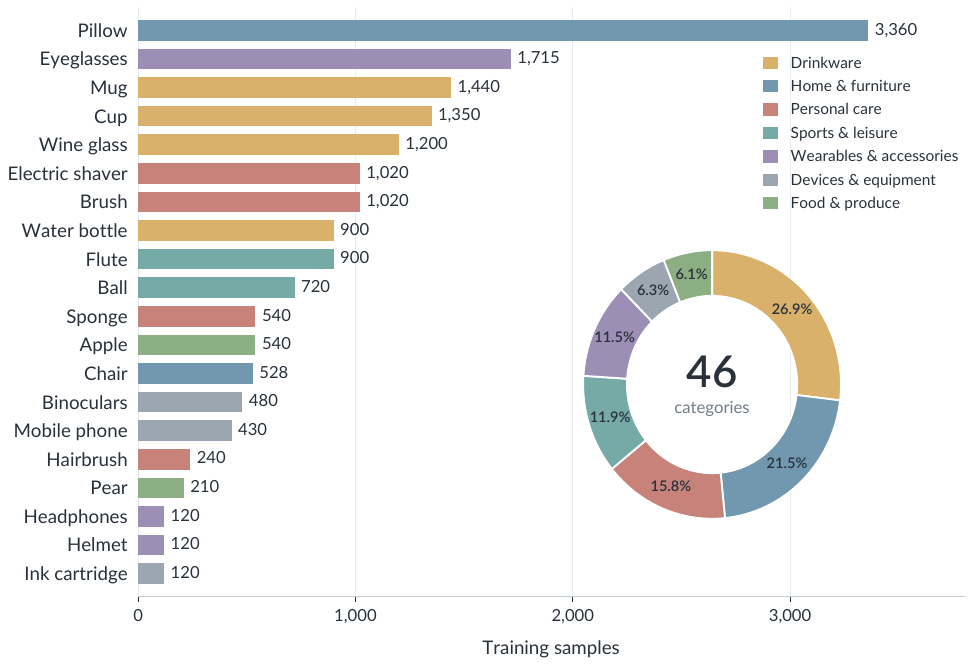}
\end{center}
\vspace{-4mm}
\caption{
\textbf{Object categories of COHE dataset.} Our COHE dataset contain head-object interactions with 46 object categories across daily objects that are frequently interacting with head. All samples, regardless of contactness, are included.}
\label{fig:supp_cohe_obj_cat}
\end{figure}

%% file: fig/supp_cohe_object_relationships.tex
\begin{figure}[t]
\begin{center}
\includegraphics[width=1.0\linewidth]{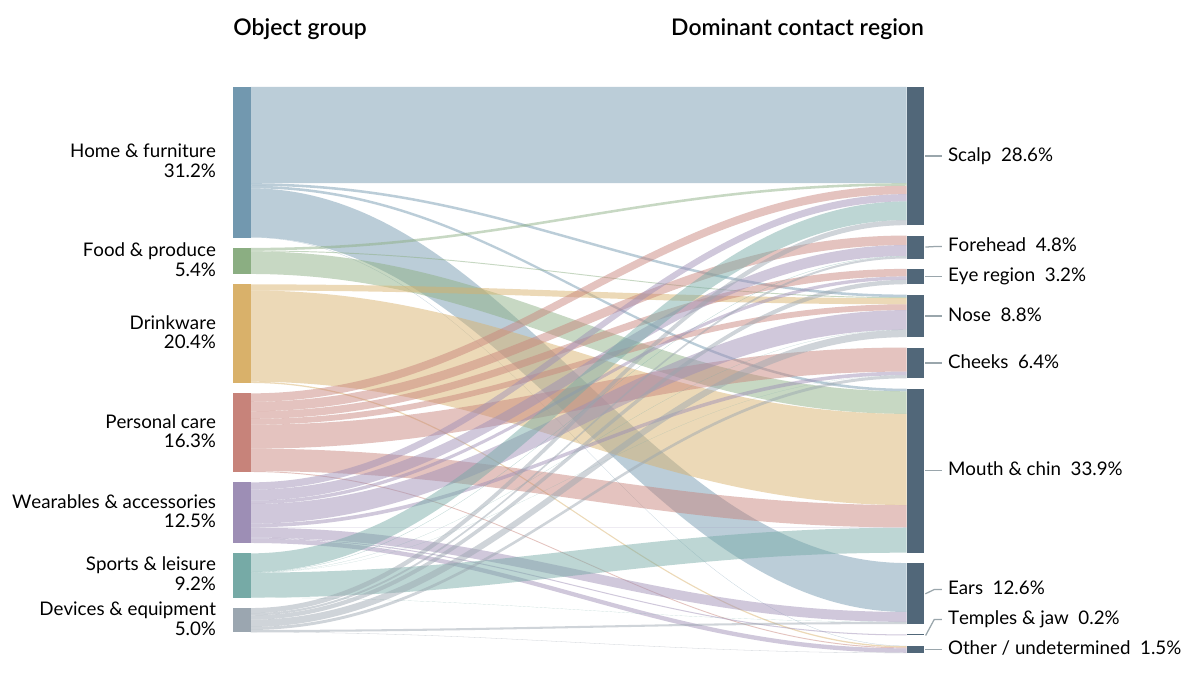}
\end{center}
\vspace{-4mm}
\caption{
\textbf{Dominant contact regions by object groups in COHE dataset.} The diagram illustrates the dominant head contact regions associated with each object group in the COHE dataset. Here, only the samples in contact are included.}
\label{fig:supp_cohe_obj_rel}
\end{figure}

%% file: tab/supp_comp_require.tex
\begin{table*}[t]
\centering
\setlength{\tabcolsep}{4pt}
\caption{\textbf{Computational requirements of various backbone configurations.}}
\scalebox{0.87}{\begin{tabular}{lcccccc} \toprule
Model & Backbone & Train Memory~(MB) & Test Memory~(MB) & Params.~(M) & Speed~(fps) & GFLOPs \\
\midrule
HECO & SigLIP2-Giant~\cite{tschannen2025siglip} & 27,554 & 12,859 & 2,069.71 & 2.84 & 1,997.56 \\
HECO & SigLIP2-So400m~\cite{tschannen2025siglip} & 22,500 & 8,155 & 1,327.02 & 4.03 & 1,120.84 \\
HECO & SigLIP2-Large~\cite{tschannen2025siglip} & 20,842 & 6,683 & 1,070.88 & 4.48 & 984.95 \\
HECO & SigLIP2-Base~\cite{tschannen2025siglip} & 17,893 & 4,175 & 561.08 & 5.08 & 711.71 \\
\bottomrule
\end{tabular}}
\vspace{-0.3cm}
\label{tab:supp_comp_require}
\end{table*}

%% file: tab/supp_sota_body_ver_cont_rep.tex
\begin{table}[t]
\centering
\footnotesize
\setlength{\tabcolsep}{2.8pt}
\renewcommand{\arraystretch}{0.9}
\caption{\textbf{Comparison with SOTA methods of dense head contact estimation in body version on DAMON~\cite{tripathi2023deco} dataset.} All methods are trained on same dataset for fair model architecture comparison.}
\vspace{-0.5em}
\begin{tabular}{@{}lccc@{}}
\toprule
Methods & Precision $\uparrow$ & Recall $\uparrow$ & F1-Score $\uparrow$ \\ 
\midrule
POSA~\cite{hassan2021populating} & 0.206 & 0.123 & 0.154 \\
BSTRO~\cite{huang2022capturing} & 0.180 & 0.172 & 0.176 \\
DECO~\cite{tripathi2023deco} & 0.167 & 0.442 & 0.243 \\
InteractVLM~\cite{dwivedi2025interactvlm} (LLaVA~\cite{liu2023visual} obj) & 0.086 & 0.377 & 0.140 \\
InteractVLM~\cite{dwivedi2025interactvlm} (GT obj) & 0.189 & 0.224 & 0.205 \\
\midrule
HECO~(Ours) & \textbf{0.381} & \textbf{0.653} & \textbf{0.481} \\
\bottomrule
\end{tabular}
\label{tab:supp_sota_body_cont_rep}
\end{table}

%% file: tab/supp_sota_head_ver_cont_rep.tex
\begin{table}[t]
\centering
\footnotesize
\setlength{\tabcolsep}{2.8pt}
\renewcommand{\arraystretch}{0.9}
\caption{\textbf{Comparison with SOTA methods of dense head contact estimation in head version on DAMON~\cite{tripathi2023deco} dataset.} All methods are trained on same dataset for fair model architecture comparison.}
\vspace{-0.5em}
\begin{tabular}{@{}lccc@{}}
\toprule
Methods & Precision $\uparrow$ & Recall $\uparrow$ & F1-Score $\uparrow$ \\ 
\midrule
POSA~\cite{hassan2021populating} & 0.149 & 0.216 & 0.176 \\
BSTRO~\cite{huang2022capturing} & 0.274 & 0.298 & 0.285 \\
DECO~\cite{tripathi2023deco} & 0.201 & 0.415 & 0.270 \\
InteractVLM~\cite{dwivedi2025interactvlm} (LLaVA~\cite{liu2023visual} obj) & 0.113 & 0.439 & 0.179 \\
InteractVLM~\cite{dwivedi2025interactvlm} (GT obj) & 0.191 & 0.479 & 0.273 \\
\midrule
HECO~(Ours) & \textbf{0.381} & \textbf{0.653} & \textbf{0.481} \\
\bottomrule
\end{tabular}
\label{tab:supp_sota_head_cont_rep}
\end{table}

%% file: tab/supp_diff_back.tex
\begin{table}[t]
\centering
\setlength{\tabcolsep}{3pt}
\caption{\textbf{Comparison of various backbone models on DAMON~\cite{tripathi2023deco} dataset.} All backbones are initialized with WebLI image--text dataset~\cite{chen2022pali}.}
\scalebox{0.85}{\begin{tabular}{lcccc} \toprule
Model & Backbone & $\text{Precision}{\uparrow}$ & $\text{Recall}{\uparrow}$ & $\text{F1-Score}{\uparrow}$ \\
\midrule
HECO & SigLIP2-Giant~\cite{tschannen2025siglip} & \textbf{0.381} & \textbf{0.653} & \textbf{0.481} \\
HECO & SigLIP2-So400m~\cite{tschannen2025siglip} & 0.348 & 0.625 & 0.447 \\
HECO & SigLIP2-Large~\cite{tschannen2025siglip} & 0.327 & 0.640 & 0.433 \\
HECO & SigLIP2-Base~\cite{tschannen2025siglip} & 0.264 & 0.633 & 0.373 \\
\bottomrule
\end{tabular}}
\label{tab:supp_diff_back}
\end{table}

%% file: fig/supp_hair_occ_amodal_inpaint_qual.tex
\begin{figure*}[t]
\begin{center}
\includegraphics[width=0.9\linewidth]{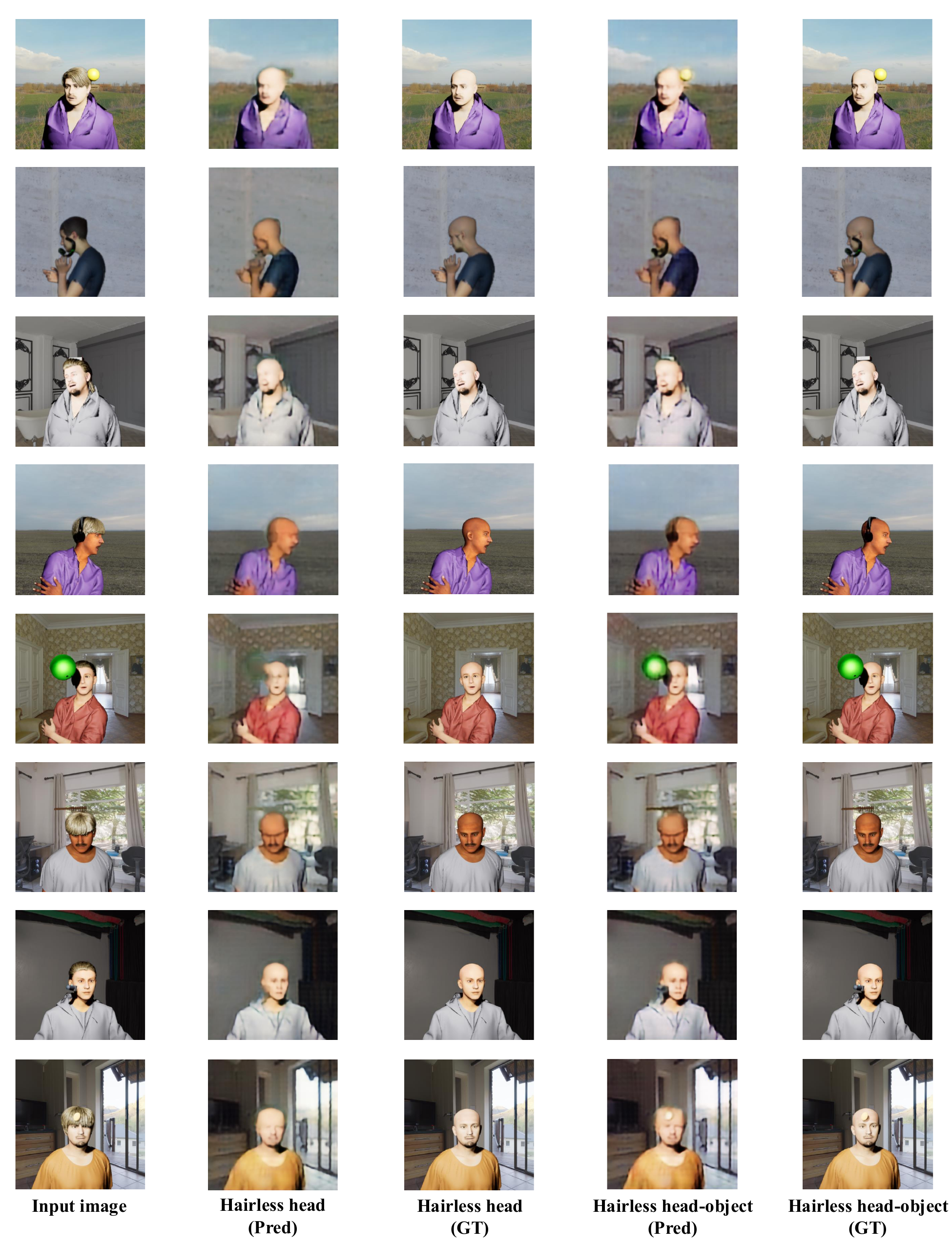}
\end{center}
\vspace{-3.5mm}
\caption{
\textbf{Visualization of amodal inpainting in hair occlusion-aware learning on COHE dataset.}}
\label{fig:supp_hair_occ_inpaint_qual}
% \vspace{-0.3cm}
\end{figure*}

%% file: fig/supp_hair_occ_amodal_mask_qual.tex
\begin{figure*}[t]
\begin{center}
\includegraphics[width=0.9\linewidth]{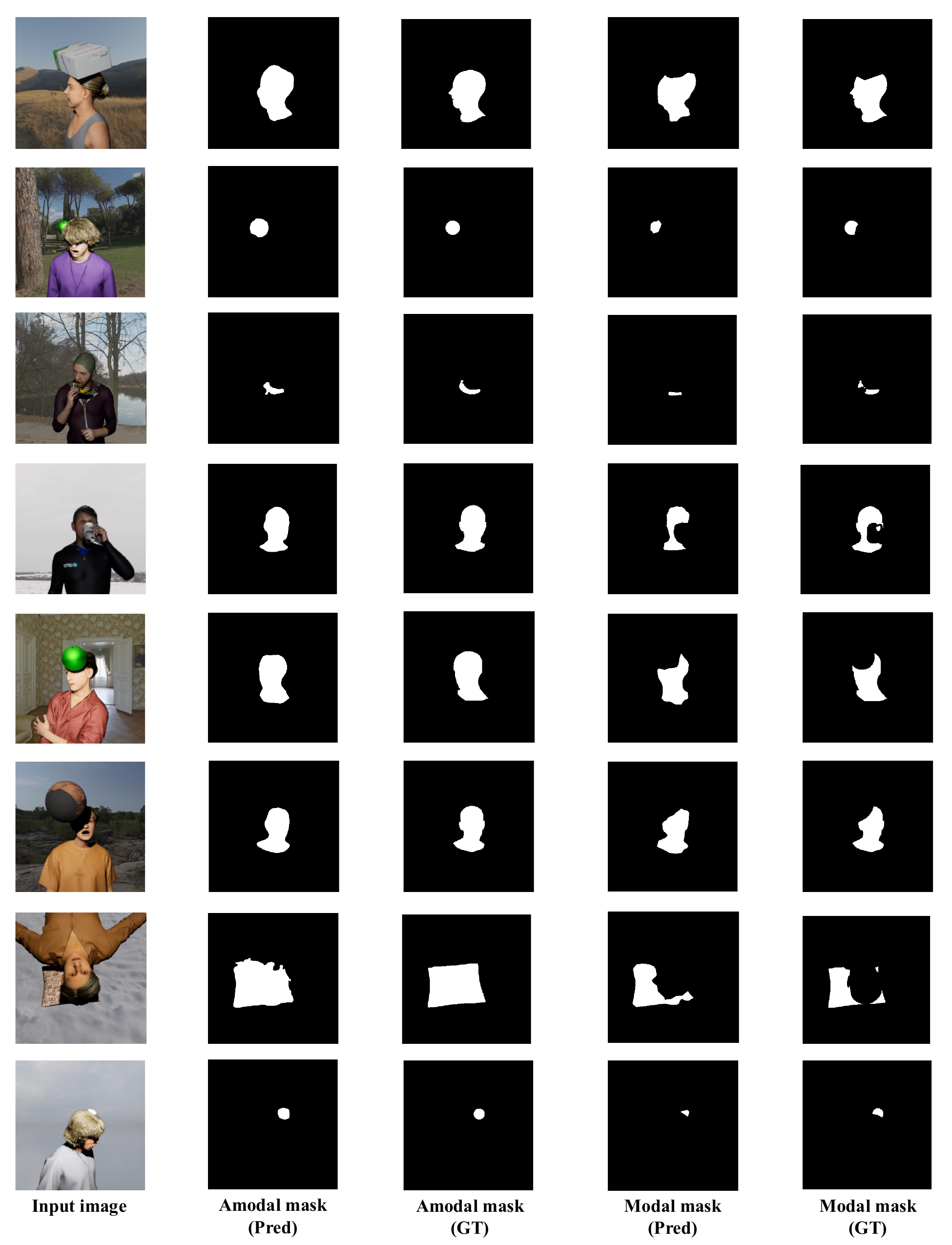}
\end{center}
\vspace{-3.5mm}
\caption{
\textbf{Visualization of amodal mask prediction in hair occlusion-aware learning on COHE dataset.}}
\label{fig:supp_hair_occ_mask_qual}
\end{figure*}

%% file: fig/supp_more_qualitative1.tex
\begin{figure*}[t]
\begin{center}
\includegraphics[width=1.0\linewidth]{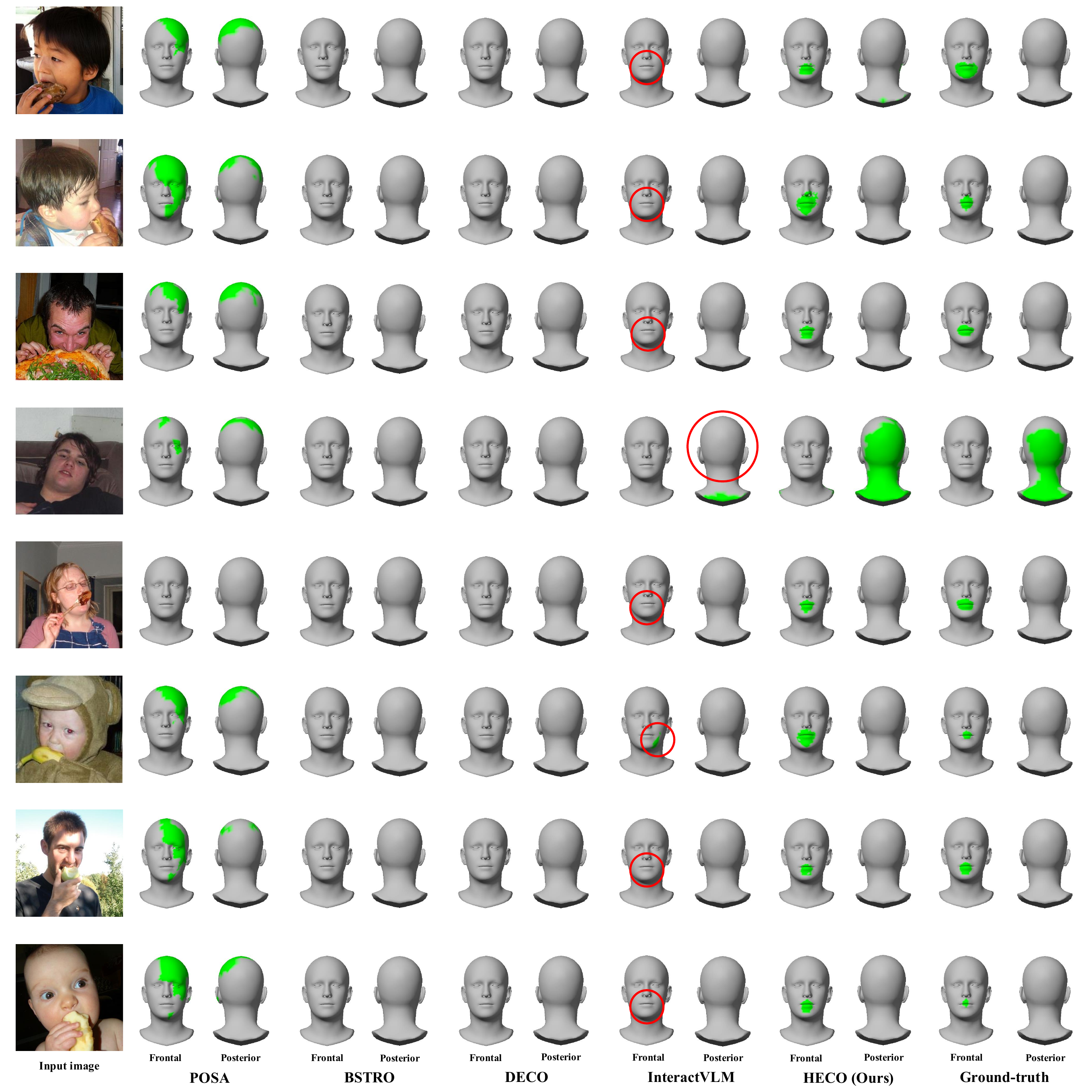}
\end{center}
\vspace{-3.5mm}
\caption{
\textbf{Qualitative comparison of dense head contact estimation with POSA~\cite{hassan2021populating}, BSTRO~\cite{huang2022capturing}, DECO~\cite{tripathi2023deco}, InteractVLM~\cite{dwivedi2025interactvlm} on DAMON~\cite{tripathi2023deco} dataset.} Red circles indicate exemplar regions that HECO outperforms previous methods.}
\label{fig:supp_more_qual1}
\end{figure*}

%% file: fig/supp_more_qualitative2.tex
\begin{figure*}[t]
\begin{center}
\includegraphics[width=1.0\linewidth]{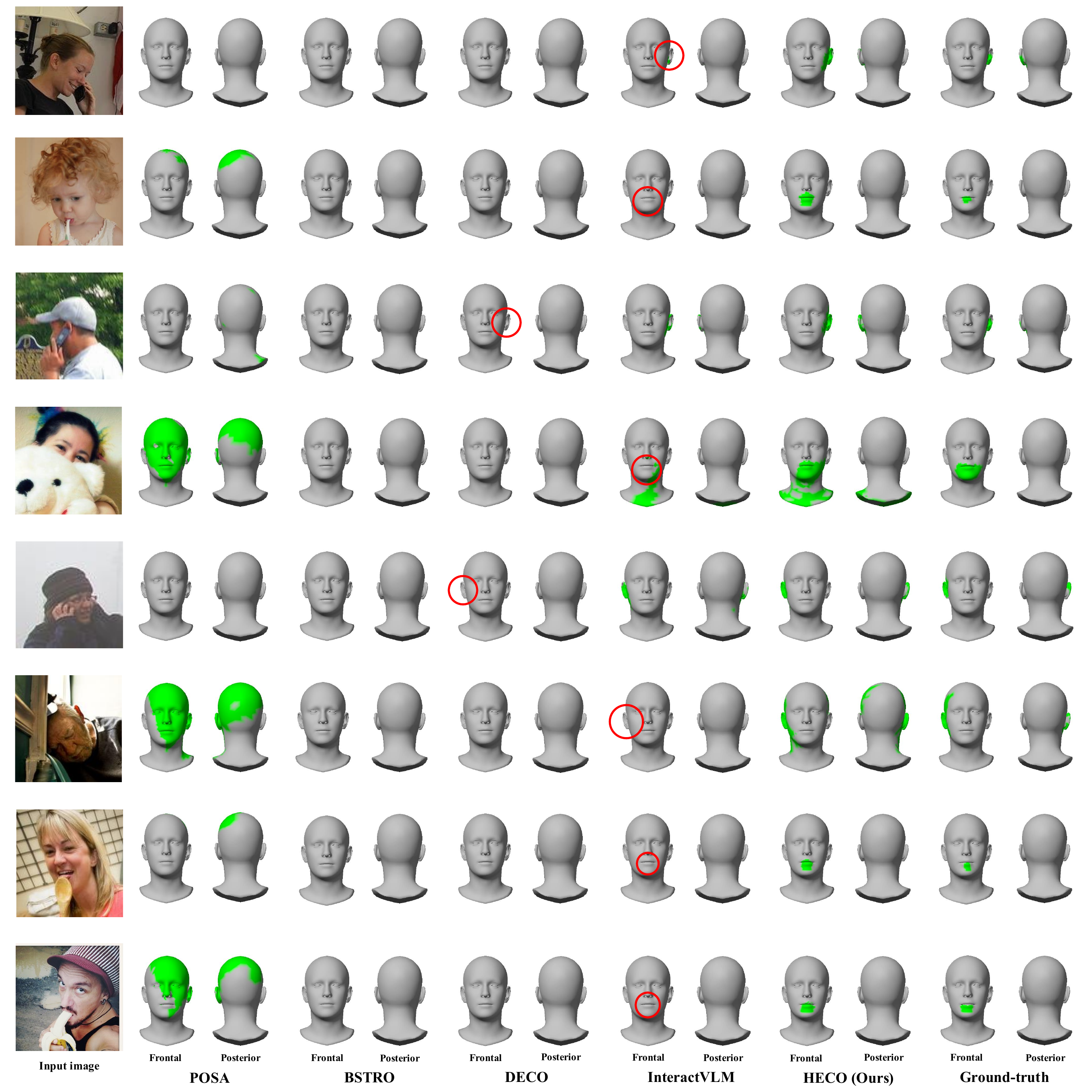}
\end{center}
\vspace{-3.5mm}
\caption{
\textbf{Qualitative comparison of dense head contact estimation with POSA~\cite{hassan2021populating}, BSTRO~\cite{huang2022capturing}, DECO~\cite{tripathi2023deco}, InteractVLM~\cite{dwivedi2025interactvlm} on DAMON~\cite{tripathi2023deco} dataset.} Red circles indicate exemplar regions that HECO outperforms previous methods.}
\label{fig:supp_more_qual2}
\end{figure*}

%% file: sec/8_llm_statement.tex
\section*{LLM Usage Statement}
In this work, we used ChatGPT solely for proofreading and improving the English grammar and phrasing of the draft. 
No LLMs were used for ideation, experimental design, code generation, or formulating scientific claims.